\documentclass{article}[11pt]

\usepackage{microtype}
\usepackage{graphicx}
\usepackage{subfigure}
\usepackage{booktabs} 

\usepackage[usenames,dvipsnames]{xcolor}
\usepackage[colorlinks=true,citecolor=blue]{hyperref}       

\usepackage[left=1.25in, top=1in, bottom=1in, right=1.25in]{geometry}
\title{Why distillation helps: a statistical perspective}
\author{Aditya Krishna Menon \and Ankit Singh Rawat \and Sashank J. Reddi \and Seungyeon Kim \qquad Sanjiv Kumar \\
Google Research, New York \\
\texttt{\{adityakmenon,ankitsrawat,sashank,seungyeonk,sanjivk\}@google.com}}

\usepackage[utf8]{inputenc} 
\usepackage[T1]{fontenc}    
\usepackage{url}            
\usepackage{booktabs}       
\usepackage{amsfonts}       
\usepackage{nicefrac}       
\usepackage{microtype}      
\usepackage{natbib}

\usepackage{float}
\usepackage{graphicx}
\usepackage[export]{adjustbox}
\usepackage[inline]{enumitem}

\usepackage{amsmath,amssymb,amsthm}
\usepackage[mathscr]{eucal}
\usepackage{stmaryrd}
\usepackage{accents}
\usepackage{upgreek}

\usepackage[most]{tcolorbox}
\usepackage{setspace}

\usepackage{booktabs,colortbl,multirow}

\usepackage{algorithm,algorithmic}

\usepackage{soul}

\newcommand{\XLR}{multiclass retrieval}
\newcommand{\pStar}{p^*}
\newcommand{\pTeach}{{p}^{\mathrm{t}}}
\newcommand{\REmp}{\hat{R}}
\newcommand{\RDist}{\tilde{R}}
\newcommand{\RDistBayes}{\hat{R}_*}

\graphicspath{{figs//}}

\allowdisplaybreaks

\newcommand{\defEq}{\stackrel{.}{=}}

\newcommand{\indicator}[1]{\llbracket #1 \rrbracket}

\renewcommand{\Pr}{\mathbb{P}}
\newcommand{\E}[2]{\underset{#1}{\mathbb{E}}\left[ #2 \right]}
\newcommand{\ES}[2]{\underset{#1}{\mathbb{E}}\, #2}

\newcommand{\X}{{x}}
\newcommand{\Y}{{y}}

\newcommand{\Z}{Z}

\newcommand{\FCal}{\mathscr{F}}
\newcommand{\HCal}{\mathscr{H}}

\newcommand{\XCal}{\mathscr{X}}
\newcommand{\YCal}{\mathscr{Y}}

\newcommand{\Real}{\mathbb{R}}

\newcommand{\tup}{\mathrm{T}}

\newtheorem{lemma}{Lemma}

\newtheorem{proposition}[lemma]{Proposition}

\theoremstyle{definition}

\newtcbtheorem[number within=section]%
{remark_tcb} 
{Remark} 
{%
fonttitle=\bfseries\upshape,fontupper=\normalsize,
colframe=green!10!black,colback=green!2!white,
colbacktitle=green!20!white,coltitle=blue!75!black,
boxsep=1pt,left=2pt,right=2pt,top=1pt,bottom=1pt,
frame hidden,boxrule=1pt,
theorem style=plain} 
{rem} 

\newtcbtheorem[number within=section]%
{assumption_tcb} 
{Assumption} 
{%
fonttitle=\bfseries\upshape,fontupper=\normalsize,
colframe=blue!5!black,colback=blue!2!white,
colbacktitle=blue!10!white,coltitle=blue!75!black,
boxsep=1pt,left=2pt,right=2pt,top=1pt,bottom=1pt,
frame hidden,boxrule=1pt,
theorem style=plain} 
{asm} 

\begin{document}

\maketitle

%

%
\begin{abstract}
Knowledge distillation is a technique for improving the performance of a simple ``student'' model by replacing its one-hot training labels with a \emph{distribution} over labels obtained from a complex ``teacher'' model. Despite proving widely effective, a basic question remains unresolved: why does distillation help? In this paper, we present a statistical perspective on distillation which 
provides one answer to this
question and suggests a novel application to multiclass retrieval.
Our core observation is that the teacher seeks to estimate the underlying (Bayes) class-probability function. Building on this,
we establish a fundamental \emph{bias-variance} tradeoff for the student: this quantifies how approximate knowledge of these class-probabilities can significantly aid learning. 
Finally,
we extend this statistical perspective to multiclass \emph{retrieval} settings,
and propose a \emph{double-distillation}
objective which
encourages a good ranking over labels.
We empirically validate our analysis, and demonstrate  the value of double-distillation for retrieval. 

\end{abstract}

\section{Introduction}
Distillation is the process of using a ``teacher'' model to improve the performance of a ``student'' model~\citep{Craven:1995,Breiman:1996,Bucilua:2006,Xue:2013,Ba:2014,Hinton:2015}.
In its simplest form, 
rather than fitting to raw labels,
one trains the student to fit the teacher's \emph{distribution} over labels.
While originally devised with the aim of model compression, 
distillation has proven successful 
in iteratively improving a fixed-capacity model
~\citep{Rusu:2016,Furlanello:2018,Yang:2019,Xie:2019},
and found use in many other settings~\citep{Papernot:2016,Tang:2016,Czarnecki:2017,Celik:2017,Yim:2017,Li:2018,Liu:2019,Nayak:2019}.

Given its empirical successes, it is natural to ask: why does distillation help?
\citet{Hinton:2015} argued that
distillation provides 
``dark knowledge'' via
teacher logits on 
the ``wrong'' labels $y' \neq y$ for an example $(x, y)$,
which effectively weights samples differently~\citep{Furlanello:2018}.
Various theoretical analyses of distillation have subsequently been developed~\citep{Lopez-Paz:2016,Phuong:2019,Foster:2019,Dong:2019,Mobahi:2020},
with particular focus on its optimisation and regularisation effects.

\begin{figure*}[!t]
    \centering
    
    \subfigure[Top-1 accuracy.]{
    \includegraphics[scale=0.45]{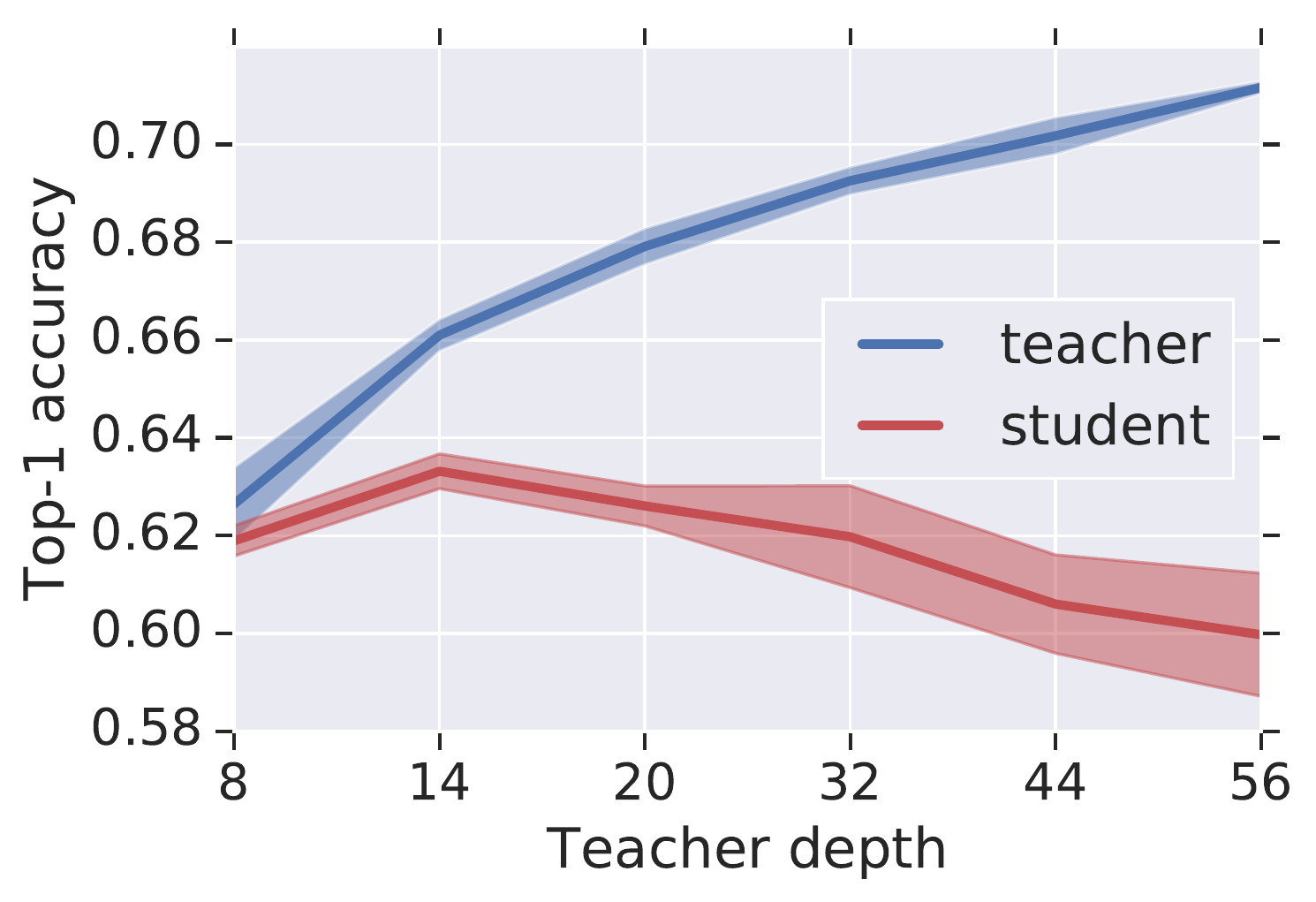}
    \label{fig:cifar100_acc}
    }%
    \subfigure[Log-loss.]{
    \includegraphics[scale=0.45]{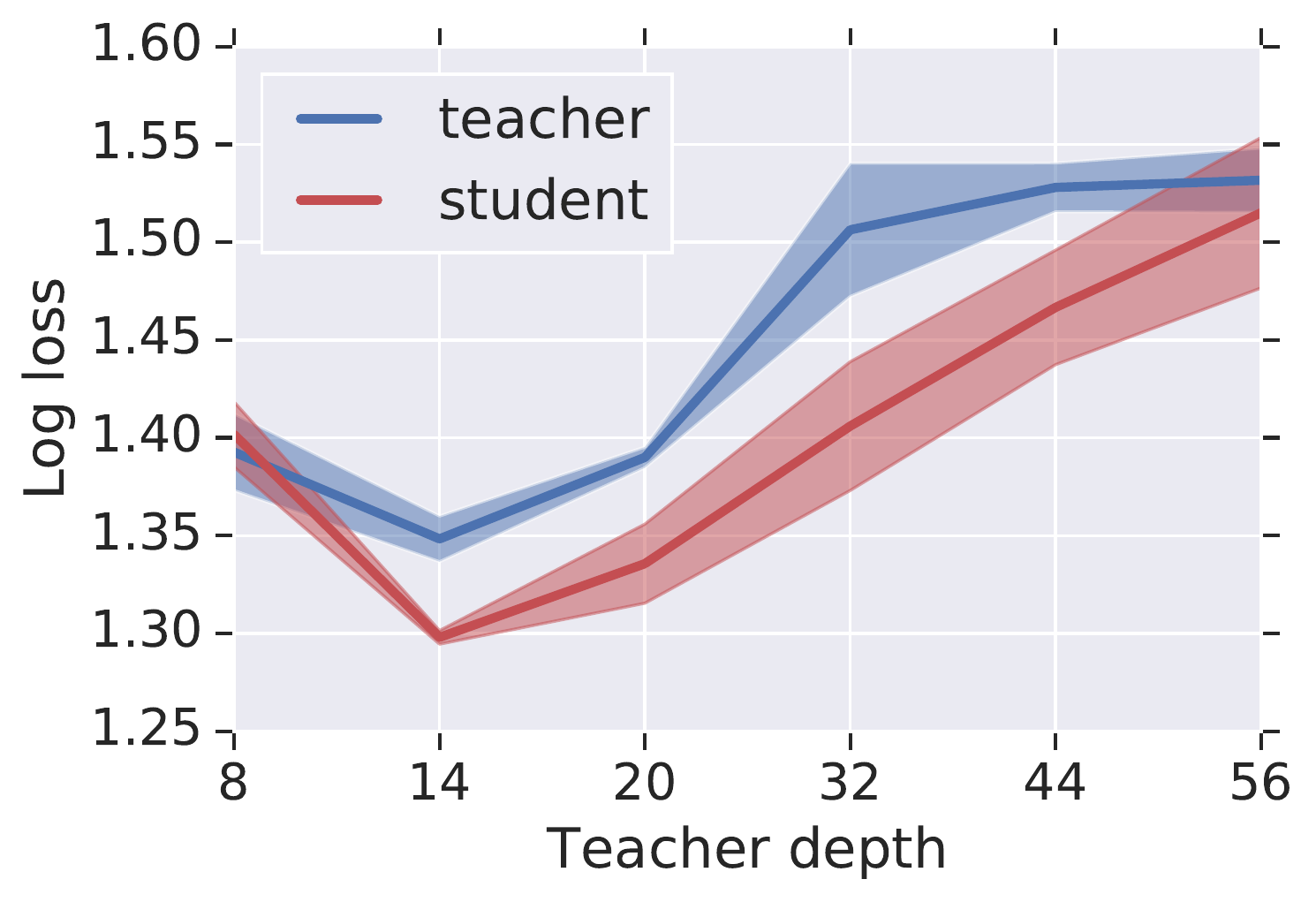}
    \label{fig:cifar100_kl}
    }  
    \subfigure[Expected calibration error.]{
    \includegraphics[scale=0.45]{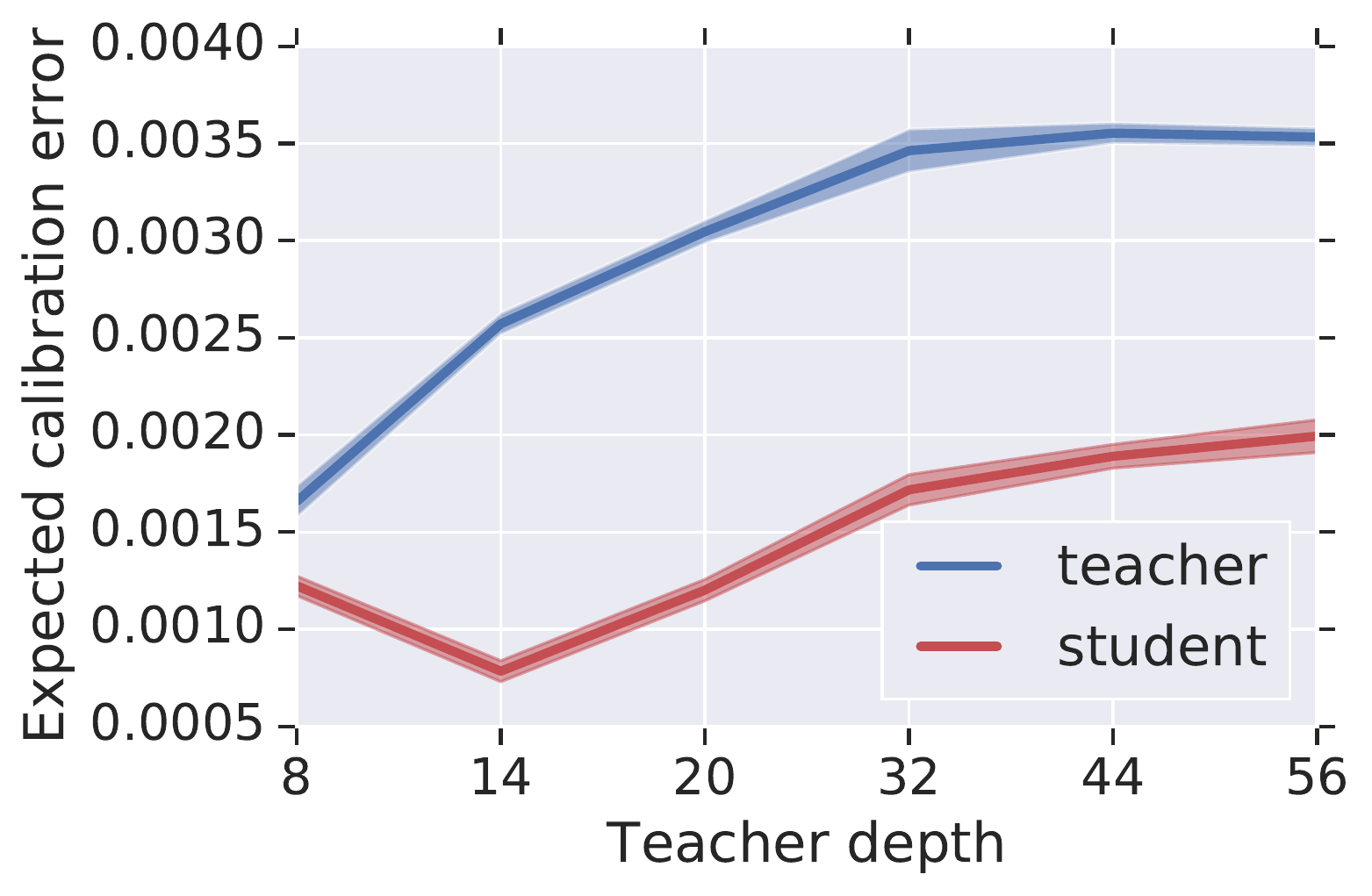}
    \label{fig:cifar100_ece}
    } 
    \caption{Illustration of 
    how better teacher modelling of underlying (Bayes) class-probabilities
    influences student generalisation,
    per our statistical perspective.
    Here, we train ResNets of varying depths on the CIFAR-100 dataset,
    and use these as teachers to distill to a student ResNet of fixed depth $8$.
    Figure~\ref{fig:cifar100_acc} reveals that the teacher model gets increasingly more accurate as its depth increases;
    \emph{however}, from~\ref{fig:cifar100_kl} 
    the corresponding log-loss
    starts increasing beyond a depth of $14$.
    The calibration error is also seen to worsen with increasing depth in~\ref{fig:cifar100_ece}.
    This indicates the teacher's probability estimates become progressively poorer approximations of the Bayes class-probability distribution $\pStar( x )$ after certain depth. Intuitively, the teacher's approximation of $\pStar$ is governed by balancing the bias and variance in its predictions.
    The accuracy of the student model also degrades beyond a teacher depth of $14$, reflecting the bound in Proposition~\ref{prop:bias-variance}. See~\S\ref{sec:bias-variance-cifar100} for more details,
    and~\S\ref{sec:bias-variance} for illustration of a general bias-variance tradeoff.
    }
    \label{fig:cifar100-distillation}
\end{figure*}

In this paper,
we present a 
novel
statistical perspective on distillation
which sheds light on why it aids performance.
Our analysis
centers on a simple observation:
\emph{a good teacher accurately models the true} (Bayes) \emph{class-probabilities}.
This is a stricter requirement than the teacher merely having high \emph{accuracy}.
We quantify how 
such probability estimates 
improve generalisation
compared to
learning from raw labels.
Building on this,
we show how
distillation 
is also useful
in 
selecting \emph{informative labels}
for
multiclass \emph{retrieval},
wherein we wish to 
order labels according to their relevance~\citep{Jain:2019}.
In sum, our contributions are:
\begin{enumerate}[label=(\roman*),itemsep=0pt,topsep=0pt]
    \item We establish the statistical benefit of using the Bayes class-probabilities in place of one-hot labels,
    and quantify a \emph{bias-variance} tradeoff
    when using approximate class-probabilities (\S\ref{sec:analysis}).

    \item We
    propose \emph{double-distillation}, 
    a novel application of distillation
    for
    \XLR{}
    wherein
    teacher probabilities guide a \emph{ranking} over labels
    (\S\ref{sec:double_distillation}).

    \item We experimentally validate 
    the value of both approximate class-probabilities in terms of generalisation,
    and of
    double-distillation
    in multiclass \emph{retrieval} (\S\ref{sec:experiments}).
\end{enumerate}
Contribution (i) gives a statistical perspective 
on the value of
``dark knowledge'': 
for an example $(x, y)$,
the logits on ``wrong'' labels $y' \neq y$ 
encode information about the underlying 
data distribution.
This view elucidates
how
a teacher's \emph{probability calibration}
rather than \emph{accuracy}
can influence a student's generalisation;
see Figure~\ref{fig:cifar100-distillation} for an illustration.
Contribution (ii) shows 
a practical benefit of
this statistical view 
of distillation,
by showing how \XLR{} objectives benefit from approximate class-probabilities.

\section{Background and notation}
We review
multiclass classification, 
retrieval, 
\&
distillation.

\subsection{Multiclass classification}

In multiclass classification,
we are given
a training sample $S \defEq \{ ( x_n, y_n ) \}_{n = 1}^N \sim \Pr^N$,
for unknown distribution $\Pr$ over instances $\XCal$ and labels $\YCal = [ L ] \defEq \{ 1, 2, \ldots, L \}$.
Our goal is to learn a predictor $f \colon \XCal \to \Real^L$ so as to minimise the \emph{risk} of $f$, i.e., its expected {loss} for a random instance and label:
\begin{equation}
    \label{eqn:risk}
    R( f ) \defEq \E{(\X, \Y) \sim \Pr}{ \ell( \Y, f( \X ) ) }.
\end{equation}
Here, $\ell \colon [ L ] \times \Real^L \to \Real_+$ is a loss function, 
where
for label $y \in [ L ]$ and prediction vector $f(x) \in \Real^L$,
$\ell( y, f(x) )$ is the loss incurred for predicting $f(x)$ when the true label is $y$.
A canonical example is
the softmax cross-entropy loss,
\begin{equation}
    \label{eqn:softmax-xent}
    \ell( y, f( x ) ) = -f_y( x ) + \log \left[ \sum_{y' \in [ L ]} e^{f_{y'}( x )} \right].
\end{equation}
We may approximate {the risk} $R( f )$ 
via
the \emph{empirical risk} 
\begin{equation}
    \label{eqn:empirical-risk-one-hot}
    \REmp( f; S ) \defEq \frac{1}{N} \sum_{n \in [N]} \ell( y_n, f( x_n ) ) 
    = \frac{1}{N} \sum_{n \in [N]} e_{y_n}^{\tup} \ell( f( x_n ) ),
\end{equation}
where 
${e}_y \in \{ 0, 1 \}^L$ denotes the one-hot encoding of $y \in [ L ]$ and 
{$\ell( f(x) ) \defEq \big(\ell(1, f(x)),\ldots, \ell(L, f(x))\big)\in \Real_+^L$} denotes the vector of losses
for each possible label.

In a multiclass \emph{retrieval} setting,
our goal is to ensure that the top-ranked labels in $f( x ) \in \Real^L$ include the true label~\citep{Lapin:2018}. 
Formally, we seek to minimise
the
\emph{top-$k$ loss}
\begin{equation}
    \label{eqn:top-k-risk}
    \ell_{\mathrm{ret}(k)}( y, f( x ) ) \defEq \indicator{y \notin \mathrm{top}_k( f( x ) )},
\end{equation}
where $\mathrm{top}_k( f )$ denotes the 
top-$k$ highest scoring labels.
When $k = 1$, this loss reduces to the standard 0-1 error.

%
\subsection{Knowledge distillation}
\label{sec:background-distillation}

Distillation involves using a ``teacher'' model to improve the performance of a  ``student'' model~\citep{Bucilua:2006,Hinton:2015}.
In the simplest form, one 
trains the teacher model and obtains
\emph{class-probability estimator} $\pTeach \colon \XCal \to \Delta_{\YCal}$,
where $\Delta$ denotes the simplex.
Each
$\pTeach_y( x )$ estimates how likely $x$ is to be classified as $y$.
In place of the empirical risk~\eqref{eqn:empirical-risk-one-hot},
the student now minimises the
\emph{distilled risk}
\begin{equation}
    \label{eqn:distillation-risk}
    \RDist( f; S ) = \frac{1}{N} \sum_{n \in [N]} \pTeach( x_n )^\tup \ell(f( x_n ) ),
\end{equation}
so that the one-hot encoding of labels is replaced with the teacher's distribution over labels.
Distillation may be used 
when the student has access to a large pool of unlabelled samples;
in this case, distillation is a means of semi-supervised learning~\citep{Radosavovic:2018}.

While originally conceived 
in settings 
where the student has lower capacity than the teacher~\citep{Bucilua:2006,Hinton:2015},
distillation has proven useful when both models have the same capacity~\citep{Breiman:1996,Furlanello:2018,Xie:2019}.
More precisely, distillation involves training a teacher on $N_t$ labelled samples $S^{\mathrm{t}} \sim \Pr^{N^{\mathrm{t}}}$,
using a function class $\FCal^{\mathrm{t}}$.
Classic distillation assumes that $N^{\mathrm{t}} \gg N$, and that the capacity of $\FCal^{\mathrm{t}}$ is greater than that of $\FCal$.
``Born-again'' distillation assumes that $S^{\mathrm{t}} = S$, and $\FCal^{\mathrm{t}} = \FCal$,
i.e.,
one iteratively trains versions of the same model, using past predictions to improve performance.

\subsection{Existing explanations for distillation}

While 
it is well-accepted that
distillation is empirically useful,
there is less consensus
as to \emph{why} this is the case.
\citet{Hinton:2015} attributed the success of distillation (at least in part) to the encoding of ``dark knowledge'' 
in the probabilities the teacher
assigns to the ``wrong'' labels $y' \neq y$ for example $(x, y)$.
This richer information plausibly aids the student,
providing for example a weighting on the samples~\citep{Furlanello:2018,Tang:2020}.
Further, 
when $\ell$ is
the softmax cross-entropy,
the gradient of the distillation objective with respect to the student logits is 
the difference in the probability estimates for the two, 
implying a form of logit matching in high-temperature regimes~\citep{Hinton:2015}.

\citet{Lopez-Paz:2016} related distillation to learning from privileged information under a noise-free setting.
~\citet{Phuong:2019} analysed the dynamics of student learning, 
assuming a deep linear model for binary classification.
~\citet{Foster:2019} provided a generalisation bound for the student, under the assumption that it learns a model close to the teacher.
This does not, however, explicate what constitutes an ``ideal'' teacher,
nor quantify how an approximation to this ideal teacher will affect the student generalisation. \citet{Gotmare:2019} studied the effect of distillation on discrimination versus feature extraction layers of the student network.
\citep{Dong:2019} argued that distillation has a similar effect to early stopping, and studied its ability to denoise labels.
Our focus, by contrast, is in settings where there is no exogenous label noise, and in studying the statistical rather than optimisation effects of distillation.
\citep{Mobahi:2020} analysed the special setting of \emph{self-distillation} -- wherein the student and teacher employ the same model class -- and showed that for kernelised models, this is equivalent to increasing the regularisation strength.

\section{Distillation through a class-probability lens}
\label{sec:analysis}
We now present a statistical perspective of distillation,
which gives insight into why it can aid generalisation.
Central to our perspective are two observations:
\begin{enumerate}[label=(\roman*),itemsep=0pt,topsep=0pt]
    \item the risk~\eqref{eqn:risk} we seek to minimise 
    inherently smooths labels by the \emph{class-probability distribution} $\Pr( y \mid x )$;
    
    \item a teacher's predictions provide an approximation to the class-probability distribution $\Pr( y \mid x )$,
    which can thus yield a better approximation to the risk~\eqref{eqn:risk}
    than one-hot labels.
\end{enumerate}
Building on these, we show how sufficiently accurate teacher approximations can improve student generalisation.

%
\subsection{Bayes knows best: distilling class-probabilities}
\label{sec:bayes-distillation}

Our starting point is the following elementary observation: 
the underlying risk $R( f )$
for a predictor $f \colon \XCal \to \Real^L$ is 
\begin{align}
\label{eq:population_risk}
    R( f ) &= \E{\X}{ \E{\Y \mid \X}{ \ell( \Y, f( \X ) ) } } \nonumber \\
    &= \E{\X}{ \pStar( \X )^{\tup} \ell( f( \X ) ) },
\end{align}
where $\pStar( x ) \defEq \begin{bmatrix} \Pr( y \mid x ) \end{bmatrix}_{y \in [ L ]} \in \Delta_{\YCal \defEq [L]}$ is the \emph{Bayes class-probability distribution}.
Intuitively, $\Pr( y \mid x )$ inherently captures the suitability of label $y$ for instance $x$.
Thus, the risk involves drawing an instance $\X \sim \Pr( {\X} )$, and then computing the average loss of $f( \X )$ over all labels $y \in [ L ]$, 
weighted by their
Bayes probabilities $\Pr( y \mid x )$.
When $\pStar( x )$ is not concentrated on a single label,
there is an \emph{inherent confusion} amongst the labels for the instance $x$.

Given an $(x_n, y_n) \sim \Pr$,
the empirical risk~\eqref{eqn:empirical-risk-one-hot}
approximates 
the distribution
$\pStar( x_n )$
with the one-hot $e_{y_n}$,
which is only supported on one label.
While this is an unbiased estimate,
it is a significant reduction in granularity. 
By contrast,
consider the following \emph{Bayes-distilled risk} on a sample $S \sim \Pr^N$:
\begin{equation}
    \label{eqn:bayes-distillation}
    \RDistBayes( f; S ) \defEq \frac{1}{N} \sum_{n = 1}^N \pStar( x_n )^{\tup} \ell( f( x_n ) ).
\end{equation}
This is a distillation objective (cf.~\eqref{eqn:distillation-risk}) using a \emph{Bayes teacher},
who provides the student with the true class-probabilities.
Rather than fitting to a single label realisation $y_n \sim \pStar( x_n )$,
a student minimising~\eqref{eqn:bayes-distillation} considers all \emph{alternate} label realisations,
weighted by their likelihood.%
\footnote{While the student could trivially memorise the \emph{training} class-probabilities,
this would not generalise to \emph{test} samples.}
Observe that when $\ell$ is the cross-entropy,~\eqref{eqn:bayes-distillation}
is simply the KL divergence between the Bayes class-probabilities and our predictions.

Both the standard 
empirical risk
$\REmp( f; S )$ in~\eqref{eqn:empirical-risk-one-hot}
and Bayes-distilled risk $\RDistBayes( f; S )$ in~\eqref{eqn:bayes-distillation} 
are unbiased estimates of the population risk $R( f )$.
But intuitively, we expect that a student minimising~\eqref{eqn:bayes-distillation}
ought to generalise better from a finite sample.
We can make this intuition precise by establishing that the Bayes-distilled risk has \emph{lower variance},
considering fresh draws of the training sample.

\begin{lemma}
\label{lemm:lower-variance}
For any fixed predictor $f \colon \XCal \to \Real^{L}$,
$$ \mathbb{V}_{S \sim \Pr^{N}}\left[ \RDistBayes( f; S ) \right] \leq \mathbb{V}_{S \sim \Pr^{N}}\left[ \REmp( f; S ) \right], $$
where $\mathbb{V}$ denotes variance,
and equality holds iff 
$\forall x \in \XCal$,
the loss values
$\ell( f( x ) )$ are constant on the support of $\pStar( x )$.
\end{lemma}

\begin{proof}[Proof of Lemma~\ref{lemm:lower-variance}]
By definition,
\begin{align*}
    \mathrm{LHS} &= \frac{1}{N} \cdot \mathbb{V}\left[ \pStar( \X )^{\tup} \ell( f( \X ) ) \right] \\
    &= \frac{1}{N} \cdot \mathbb{V}\left[ \E{\Y \mid \X}{ \ell( \Y, f( \X ) ) } \right] \\
    &= \frac{1}{N} \cdot \ES{\X}{ \left[ \E{\Y \mid \X}{ \ell( \Y, f( \X ) ) } \right]^2 } -
    \frac{1}{N} \cdot \left[ \ES{\X}{\E{\Y \mid \X}{ \ell( \Y, f( \X ) ) }} \right]^2.
\end{align*}
\begin{align*}
    \mathrm{RHS} &= \frac{1}{N} \cdot \mathbb{V}\left[ \ell( \Y, f( \X ) ) \right] \\
    &= \frac{1}{N} \cdot \ES{\X}{\E{\Y \mid \X}{ \ell( \Y, f( \X ) )^2 }} - \frac{1}{N} \cdot \left[ \ES{\X}{\E{\Y \mid \X}{ \ell( \Y, f( \X ) ) }} \right]^2.
\end{align*}
In both cases, the second term simply equals $R( f )^2$
since both estimates are unbiased.
For fixed $x \in \XCal$, 
the result follows by Jensen's inequality applied to the random variable 
$\Z \defEq \ell( \Y, f( x ) )$.
Equality occurs iff $\Z$ is constant,
which requires the loss to be constant on the support of $\pStar( x )$.
\end{proof}

The condition when the Bayes-distilled and empirical risks have the same variance is intuitive:
the two risks trivially agree
when $f$ is non-discriminative (attaining equal loss on all labels),
or when a label is inherently deterministic (the class-probability is concentrated on one label).
For discriminative predictors and non-deterministic labels,
however,
the Bayes-distilled risk can have significantly lower variance.

The reward of reducing variance is better generalisation:
a student minimising~\eqref{eqn:bayes-distillation},
will better minimise the population risk,
compared to using one-hot labels.
In fact, we may quantify how the \emph{empirical} variance of the Bayes-distilled loss values
influences generalisation as follows.

\begin{proposition}
\label{prop:svp}
Pick any bounded loss $\ell$.
Fix a hypothesis class $\FCal$ of predictors 
$f \colon \XCal \to \Real^L$,
with induced class $\HCal^* \subset [ 0, 1 ]^{\XCal}$ 
of functions $h( x ) \defEq \pStar( x )^{\tup} \ell( f( x ) )$.
Suppose $\HCal^*$
has uniform covering number 
$\mathscr{N}_{\infty}$.
Then,
for any $\delta \in (0, 1)$,
with probability at least $1 - \delta$ over $S \sim \Pr^N$,
$$ R( f ) \leq \RDistBayes( f; S ) + \mathscr{O}\left( \sqrt{ \mathbb{V}^*_N( f ) \cdot \frac{\log \frac{\mathscr{M}^*_N}{\delta}}{N}} + \frac{\log \frac{\mathscr{M}^*_N}{\delta}}{N} \right), $$
where $\mathscr{M}^*_N \defEq \mathscr{N}_{\infty}( \frac{1}{N}, \HCal^*, 2 N )$
and $\mathbb{V}^*_N( f )$ is the empirical variance of the loss values $\{ \pStar( x_n )^{\tup} \ell( f( x_n ) ) \}_{n = 1}^N$.
\end{proposition}

\begin{proof}[Proof of Proposition~\ref{prop:svp}]
This a simple consequence of~\citet[Theorem 6]{Maurer:2009},
which is a uniform convergence version of Bennet's inequality~\citep{Bennett:1962}.
\end{proof}

The above may be contrast to the
bound achievable for the standard empirical risk using one-hot labels:
by~\citet[Theorem 6]{Maurer:2009},
$$ R( f ) \leq \REmp( f; S ) + \mathscr{O}\left( \sqrt{ \mathbb{V}_N( f ) \cdot \frac{\log \frac{\mathscr{M}_N}{\delta}}{N}} + \frac{\log \frac{\mathscr{M}_N}{\delta}}{N} \right), $$
where here we consider a function class $\HCal$ comprising functions $h( x, y ) = e_{y}^{\tup} \ell( f( x ) )$,
with uniform covering number $\mathscr{M}_N$.
Combining the above with Lemma~\ref{lemm:lower-variance}, we see that the Bayes-distilled empirical risk results in a lower variance penalty.%

%
To summarise the statistical perspective
espoused above,
a student should ideally have access to the underlying class-probabilities $\pStar( x )$,
rather than a single realisation $y \sim \pStar( x )$.
As a final comment, the above provides
a statistical perspective on the value of ``dark knowledge'':
the teacher's ``wrong'' logits $y' \neq y$ for an example $(x, y)$
provide approximate information about the Bayes class-probabilities.
This results in a lower-variance student objective, 
aiding generalisation.

%
\subsection{Distilling from an imperfect teacher}
\label{sec:bias-variance}

The previous section explicates how an idealised ``Bayes teacher''
can be beneficial to a student.
How does this translate to more realistic settings, where one obtains predictions from a teacher which itself is learned form data?

Our first observation is that
a teacher's predictor $\pTeach$ can typically be seen as an 
imperfect estimate of the true $\pStar$.
Indeed, if the teacher is trained with a loss that is \emph{proper}~\citep{Savage:1971,Schervish:1989,Buja:2005}, 
then
\begin{equation}
    \label{eqn:risk-as-divergence}
    R( f ) = \E{\X}{ D( \pStar( \X ) \| \pTeach( \X ) ) } + \mathrm{constant},
\end{equation}
where $D( \cdot \| \cdot )$ is a (loss-dependent) \emph{divergence} between the true and estimated class-probability functions.
For example, the softmax cross-entropy corresponds to $D( \cdot \| \cdot )$ being the KL divergence between the distributions.
The teacher's goal is thus fundamentally to ensure that their predictions align with the true class-probability function.

Of course,
a teacher learned from finite samples is unlikely to achieve zero divergence in~\eqref{eqn:risk-as-divergence}.
Indeed, even high-capacity teacher models may not be rich enough capture the true $\pStar$.
Further, even if the teacher can represent $\pStar$,
it may not be able to \emph{learn} this perfectly given a finite sample,
owing to both 
statistical (e.g., the risk of overfitting) and 
optimisation (e.g., non-convexity of the objective) issues.
We must thus treat $\pTeach$ as an \emph{imperfect} estimate of $\pStar$.
The natural question is:
will such an estimate still improve generalisation?

To answer this,
we establish a fundamental \emph{bias-variance} tradeoff when performing distillation.
Specifically, we show the difference between the distilled risk $\RDist( f; S )$ (cf.~\eqref{eqn:distillation-risk}) and population risks $R( f )$ (cf.~\eqref{eq:population_risk})
depends on how variable the loss under the teacher is,
and how well the teacher estimates $\pStar$ in a squared-error sense. Intuitively, the latter captures how well the teacher estimates $\pStar$ on average (bias),
and how variable the teacher's predictions are (variance).
\begin{proposition}
\label{prop:bias-variance}
Pick any bounded loss $\ell$.
Suppose we have a teacher model $\pTeach$ with corresponding distilled empirical risk in \eqref{eqn:distillation-risk}.
For constant $C > 0$
and any predictor $f \colon \XCal \to \Real^L$,
\begin{align}
\label{eq:bias-variance-1}
\E{}{ ( \RDist( f; S ) - R( f ) )^2 } &\leq \frac{1}{N} \cdot \mathbb{V}\left[  \pTeach( \X )^{\tup} \ell( f( \X ) ) \right] + C \cdot \left( \E{}{ \| \pTeach( \X ) - \pStar( \X ) \|_2 } \right)^2 \\
&\leq \frac{1}{N} \cdot \mathbb{V}\left[  \pTeach( \X )^{\tup} \ell( f( \X ) ) \right] + C \cdot \left( { \| \E{}{ \pTeach( \X ) } - \pStar( \X ) \|_2^2 } + \mathbb{V}\left[ \pTeach( \X ) \right] \right), \label{eq:bias-variance-2}
\end{align}
where $\mathbb{V}\left[ \cdot \right]$ denotes the sum of coordinate-wise variance.
\end{proposition}

\begin{proof}[Proof of Proposition~\ref{prop:bias-variance}]
Let $\Delta \defEq \RDist( f; S ) - R( f )$. 
Then,
\begin{align*}
    \E{}{ ( \RDist( f; S ) - R( f ) )^2 } &= \E{}{ \Delta^2 } = \mathbb{V}[ \Delta ] + \E{}{ \Delta }^2.
\end{align*}
Observe that
\begin{align*}
    \E{}{ \Delta } &= \E{\X}{ (\pTeach( \X ) - \pStar( \X ))^{\tup} \ell( f( \X ) ) } \\
    &\leq \E{\X}{ \| \pTeach( \X ) - \pStar( \X ) \|_2 \cdot \| \ell( f( \X ) ) \|_2 } \\
    &\leq \E{\X}{ \| \pTeach( \X ) - \pStar( \X ) \|_2 \cdot C \cdot \| \ell( f( \X ) ) \|_{\infty} } \\
    &\leq C \cdot \E{\X}{ \| \pTeach( \X ) - \pStar( \X ) \|_2 },
\end{align*}
where the second line is by the Cauchy-Schwartz inequality,
and the third line by the equivalence of norms.
Now, \eqref{eq:bias-variance-1} follows since $R( f )$ is a constant, implying
\begin{align*}
\mathbb{V}[ \Delta ] &= \mathbb{V}[ \RDist( f; S ) ] = \frac{1}{N} \cdot \mathbb{V}\left[  \pTeach( \X )^{\tup} \ell( f( \X ) ) \right].
\end{align*}
For \eqref{eq:bias-variance-2}, by Jensen's inequality,
and the definition of variance,
\begin{align*}
    \left( \E{\X}{ \| \pTeach( \X ) - \pStar( \X ) \|_2 } \right)^2 &\leq \E{\X}{ \| \pTeach( \X ) - \pStar( \X ) \|_2^2 } \\
    &= \| \E{}{ \pTeach( \X ) } - \pStar( \X ) \|_2^2 + \mathbb{V}\left[ \pTeach( \X ) \right].
\end{align*}
\end{proof}

Unpacking the above, the fidelity of the distilled risk's approximation to the true one depends on three factors:
how variable the expected loss is for a random instance; how well the teacher's $\pTeach$ approximates the true $\pStar$ on average; and how variable the teacher's predictions are.
Mirroring the previous section,
we may convert Proposition~\ref{prop:bias-variance}
into a generalisation bound for the student:
\begin{equation}
    \label{eqn:bias-variance-generalisation}
    R( f ) \leq \RDist( f; S ) + \mathfrak{C}( \FCal, N ) + \mathscr{O}\left(\mathbb{E}{}{ \| \pTeach( \X ) - \pStar( \X ) \|_2}\right),
\end{equation}
where 
$\mathfrak{C}$ is the penalty term from Proposition~\ref{prop:svp}.
As is intuitive, using an imperfect teacher invokes an additional penalty depending on how far the predictions are from the Bayes, in a squared-error sense. For completeness, a formal statement is provided in Proposition~\ref{prop:svp-phat} in Appendix~\ref{sec:appendix-theory}.

\subsection{Discussion and implications}

Our statistical perspective gives a simple yet powerful means of understanding distillation.
Our formal results follow readily from this perspective,
but their implications are subtle, and merit further discussion.

%
$\triangleright$ \emph{Why accuracy is not enough}.
Our bias-variance result 
establishes that if the teacher provides good class-probabilities,
in the sense of approximating $\pStar$ in a mean-square sense,
then the resulting student should generalise well.
In deep networks,
this does
\emph{not} imply providing the teacher having higher accuracy;
such models
may be accurate while being overly confident%
~\citep{Guo:2017,Rothfuss:2019}.
Our result thus potentially illuminates why more accurate teachers may lead to poorer students,
as has been noted empirically~\citep{Muller:2019};
see also~\S\ref{sec:bias-variance-cifar100}. 

In practice,
the precise bound derived above is expected to be loose.
However, its qualitative trend may be observed in practice.
Figure~\ref{fig:cifar100-distillation} (see also~\S\ref{sec:bias-variance-cifar100}) illustrates 
how increasing the depth of a ResNet model may increase accuracy, but degrade probabilistic calibration.
This is seen to directly relate to the quality of a student distilled from these models.

Temperature scaling~\citep{Hinton:2015}, a very common and empirically successful trick in distillation, 
can also be analysed through this perspective. 
Usually, teachers are highly complex and optimised to maximise accuracy; hence, they often become overly confident. 
Increasing the temperature can help the student-target 
be closer to the true distribution, 
and not just convey the most accurate label.

$\triangleright$ \emph{Teacher variance and model capacity}.
Proposition~\ref{prop:bias-variance} allows for $\pTeach$ to be random (e.g., owing to the teacher being learned from some independent sample $S^{\mathrm{t}} \sim \Pr^{N^{\mathrm{t}}}$).
Consequently, 
the variance terms not only reflect how diffused the teacher's predictions are,
{but}
also how much these predictions \emph{vary} over fresh draws of the \emph{teacher} sample.
\emph{High-capacity} teachers may yield vastly different predictions when trained on fresh samples.
This variability incurs a penalty in the third term in Proposition~\ref{prop:bias-variance}.
At the same time, such teachers can better estimate the Bayes $\pStar( x )$,
which incurs a lower bias.
The delicate tradeoff between these concerns translates into (student) generalisation.

We may understand
the 
\emph{label smoothing} trick~\citep{Szegedy:2016}
in light of the above.
This corresponds to 
mixing the student labels with uniform predictions, yielding
$\pTeach( x ) = (1 - \alpha) \cdot e_{y} + \frac{\alpha}{L} \cdot \mathbf{1}$
for $\alpha \in [ 0, 1 ]$.
From the perspective of modelling $\pStar$, 
choosing $\alpha > 0 $ introduces a \emph{bias}.
{However},
$\pTeach( x )$ has lower \emph{variance} than the one-hot labels $e_y$'s,
owing to the $(1 - \alpha)$ scaling.
Provided the bias is not too large, 
smoothing can thus aid generalisation.

We remark also that the first variance term in Proposition~\ref{prop:bias-variance} 
vanishes as $N \to \infty$.
This is intuitive: in the limit of infinite student samples,
the quality of the distilled objective is wholly determined by how well the teacher probabilities model the Bayes probabilities.
For small $N$, 
this term measures how diffused the losses are when weighted by the teacher probabilities (similar to Lemma~\ref{lemm:lower-variance}).

$\triangleright$ \emph{How much teacher bias is admissible?}
When $\pTeach \neq \pStar$,
Proposition~\ref{prop:bias-variance} 
reveals that a distilled student's generalisation gap depends on the bias and variance in $\pTeach$.
By contrast,
from a labelled sample, 
the student's generalisation gap depends on the complexity of its model class $\FCal$.
Distillation can be expected to help when the first gap is lower than the second.

Concretely, 
when trained from limited labelled samples, the student will find
$\min_{f \in \FCal} \hat{R}( f; S )$,
which incurs a high \emph{statistical} error.
Now
suppose we have a large amount of unlabelled data $N \to \infty$.
A distilled student can then reliably find the minimiser (following~\eqref{eqn:risk-as-divergence})
$$ \min_{f \in \FCal} \E{\X}{ D( \pTeach( \X ) \| f( \X )  ) }, $$
with essentially no statistical error,
but an \emph{approximation} error given by the teacher's bias.
In this setting, we thus need
\emph{the teacher's bias to be lower than the statistical error}.

$\triangleright$ \emph{On teacher versus student samples}
A qualifier to our results is that they assume a disjoint set of samples for the teacher and student.
For example, it may be that the teacher is trained on a pool of labelled samples,
while the student is trained on a larger pool of unlabelled samples.
The results thus do not directly hold for the settings of \emph{self-} or \emph{co-}distillation~\citep{Furlanello:2018,Anil:2018},
wherein the same sample is used for both models.
Combining our results with recent analyses
from the perspective of regularisation~\citep{Mobahi:2020} and 
data-dependent function classes~\citep{Foster:2019} would be of interest in future work.

\section{Distillation meets multiclass retrieval}
\label{sec:double_distillation}
Our statistical view has thus far given insight into the potential value of distillation.
We now 
show a distinct practical benefit of this view, 
by leveraging it for a novel application of distillation to
\emph{multiclass retrieval}.
Our basic idea is to 
construct a
\emph{double-distillation} objective, 
wherein distillation informs the loss as to the relative confidence of both ``positive'' \emph{and} ``negative'' labels.

%
\subsection{Double-distillation for multiclass retrieval}

Recall from~\eqref{eqn:top-k-risk}
that in multiclass retrieval,
we wish to ensure that the top-ranked labels in our predictor $f( x ) \in \Real^L$ contain the true label.
The softmax-cross entropy $\ell( y, f( x ) )$ in~\eqref{eqn:softmax-xent}
offers a reasonable surrogate loss for this task, since
$$ \lim_{\gamma \to 0} \frac{1}{\gamma} \cdot \log\left[ \sum_{y' \in [L]} e^{\gamma \cdot (f_{y'}( x ) - f_{y}( x ))} \right] = \max_{y' \neq y} f_{y'}( x ) - f_{y}( x ). $$
The latter is 
related to
the Cramer-Singer loss~\citep{Crammer:2002},
which bounds the top-$1$ retrieval loss.
Given a sample $\{ ( x_n, y_n ) \}_{n = 1}^N$,
we thus seek to minimise $\ell( y_n, f( x_n ) )$.
From our statistical view,
there is value in instead computing $\pStar( x_n )^{\tup} \ell( f( x_n ) )$:
intuitively,
each $y \in [L]$ acts a ``smooth positive'',
weighted by the Bayes probability $\pStar_y( x_n )$.

We observe,
however,
that such smoothing does \emph{not} affect the innards of the loss itself.
In particular,
one critique of~\eqref{eqn:softmax-xent} is that it assigns equal importance to each $y' \in [ L ]$.
Intuitively,
mistakes on labels $y'$ that are a \emph{poor} explanation for $x$ 
---
i.e., have \emph{low} $\pStar_{y'}( x )$
---
ought to be strongly penalised.
On the other hand, if under the Bayes probabilities some $y' \neq y$ strongly explains $x$,
it ought to be ignored.

To this end, 
consider a generalised softmax cross-entropy:
\begin{equation}
    \label{eqn:generalised-xent}
    \ell( y, f( x ) ) \defEq \log \E{\Y' \sim P( x )}{ e^{f_{\Y'}( \X ) - f_{y}( \X )} },
\end{equation}
where $P( x ) \in \Delta_L$ is a distribution over labels.
When $P( x )$ is uniform, this is exactly the standard cross-entropy loss~(cf.~\eqref{eqn:softmax-xent}) plus a constant.
For non-uniform $P( x )$, however, the loss 
is encouraged to focus on ensuring
$f_{y}( x ) \gg f_{y'}( x )$
for those $y'$ with large $P_{y'}( x )$. 

%
Continuing our statistical view, we posit that an ideal choice of $P$
is $P( x ) = \Psi( \pStar( x ) )$,
where $\Psi( \cdot )$ is some decreasing function.
Concretely,
our new loss for a given $(x, y)$ is
$$ \ell( y, f( x ) ) = \log \sum_{y' \in [L]} \Psi( \pStar_{y'}( x ) ) \cdot e^{f_{y'}( x ) - f_{y}( x )}. $$
Intuitively, 
the loss treats $y$ as a ``positive'' label for $x$,
and each $y' \in [L]$ where $y' \neq y$ as a ``negative'' label.
The loss seeks to score the positive label over 
\emph{all labels which poorly explain $x$}.
Compared to the standard softmax cross-entropy,
we avoid penalisation when $y'$ plausibly explains $x$ as well.

Recall that under a distillation setup, a teacher model provides us estimates $\pTeach( x )$ of $\pStar( x )$.
Consequently, 
to estimate this risk on a finite sample $S = \{ ( x_n, y_n ) \}_{n = 1}^N$,
we may construct
the following \emph{double-distillation} objective:
\begin{equation}
    \label{eqn:double-distillation}
    \begin{aligned}
    \RDist( f; S ) &= \frac{1}{N} \sum_{n = 1}^{N} \pTeach( x_n )^{\tup} \ell( f( x_n ) ) \\
    \ell( y, f( x_n ) ) &= \log \left[ \overline{\pTeach}( x_n )^{\tup} e^{f( x_n ) -  f_{y}( x_n )} \right],
    \end{aligned}
\end{equation}
where $\overline{\pTeach}( x ) \defEq \Psi( {\pTeach}( x ) )$;
unrolling,
this corresponds to the objective
$$ \RDist( f; S ) = \frac{1}{N} \sum_{n = 1}^{N} \sum_{y \in [L]} \pTeach( y \mid x_n ) \cdot \log \left[ \sum_{y' \in [L]} \overline{\pTeach}( y' \mid x_n ) \cdot e^{f_{y'}( x_n ) -  f_{y}( x_n )} \right]. $$
Observe that~\eqref{eqn:double-distillation} uses the teacher in two ways: the first is the standard use of distillation to smooth the ``positive'' training labels.
The second is a novel use of distillation to smooth the \emph{``negative''} labels.
Here, for any candidate ``positive'' label $y$,
we apply varying weights to ``negative'' labels $y' \neq y$
when computing 
each
$\ell( y, f( x ) )$.

It remains to specify the precise form of $\Psi( \cdot )$ in~\eqref{eqn:double-distillation}.
A natural choice of $1 - \pTeach( x )$;
however, since the entries of $\pTeach( x )$ sum to one,
this may not allow the loss to sufficiently ignore other ``positive'' labels.
Concretely, suppose there are $L^+$ plausible ``positive'' labels,
which have most of the probability mass under $\pTeach( x )$.
Then, the total weight assigned to these labels will be roughly $L^+ - 1$;
this is of the order of $L^+$, which is what we would get from uniform weights.

To resolve this, we may instead use weights 
proportional to
$1 - \sigma( a \cdot s_{y}( x ) )$,
where $\sigma( \cdot )$ is the sigmoid function,
$s_y( x )$ is the teacher \emph{logit},
and $a > 0$ is a scaling parameter which may be tuned.
This parameterisation allows for multiple labels to have high (or low) weights.
In the above example,
each ``positive'' label can individually get a score close to $0$.
The total weight of the ``positives'' can thus also be close to $0$,
and so the loss can learn to ignore them.

\subsection{Discussion and implications}

The viability of distillation as a means of smoothing negatives has not, to our knowledge, been explored in the literature.
The proposal is, however, a natural consequence of our statistical view of distillation developed in~\S\ref{sec:analysis}.

The double-distillation objective in \eqref{eqn:double-distillation} relates to ranking losses.
In bipartite ranking settings~\citep{Cohen:1999},
one assumes an instance space $\XCal$ and binary labels $\YCal = \{ 0, 1 \}$ denoting that items are either relevant, or irrelevant.
\citet{Rudin:2009} proposed the following \emph{push loss} for this task:
\begin{align*}
    R_{\mathrm{push}}( h ) &\defEq \E{z \sim \Pr_{+}}{ g\left( \E{z' \sim \Pr_{-}}{ \phi( h( z' ) - h( z ) ) } \right) }
\end{align*}
where $\Pr_{+}$ and $\Pr_{-}$ are distributions over positive and negative instances respectively,
and $g( \cdot ), \phi( \cdot )$ are convex non-decreasing functions.
Similar losses have also been studied in~\citet{Yun:2014}.
The generalised softmax cross-entropy in~\eqref{eqn:generalised-xent}
can be seen as a \emph{contextual} version of this objective for each $x \in \XCal$,
where $g( z ) = \log( 1 + z )$ and $\phi( z ) = e^{-z}$.
In double-distillation, 
the contextual distributions $\Pr_{\pm}( x )$ are approximated using the teacher's predictions $\pTeach( x )$.

There is a broad literature on using a weight on ``negatives'' in the softmax~\citep{Liu:2016,Liu:2017,Wang:2018,Li:2019,Cao:2019};
this is typically motivated by ensuring a varying margin for different classes.
The resulting weighting is thus either constant or label-dependent,
rather than the label- and example-dependent weights provided by distillation.
Closer still to our framework is the recent work of~\citet{Khan:2019},
which employs uncertainty estimates in the predictions for a given example and label to adjust the desired margin in the softmax.
While not explicitly couched in terms of distillation,
this may be understood as a ``self-distillation'' setup,
wherein the current predictions of a model are used to progressively refine future iterates.
Compared to double-distillation, however, the nature of the weighting employed is considerably more complicated.



There is a rich literature on the problem of label ranking,
where typically it is assumed that one observes a (partial) ground-truth ranking over labels~\citep{Dekel:2004,Furnkranz:2008,Vembu:2011}.
We remark also that the 
view of
the softmax as a ranking loss has received recent attention~\citep{Bruch:2019,Bruch:2019b}.
Exploiting the statistical view of distillation in these regimes is a promising future direction.
\citet{Tang:2018} explored distillation in a related learning-to-rank framework.
While similar in spirit,
this focusses on pointwise losses, 
wherein the distinction between positive and negative smoothing is absent.

Finally, we note that while our discussion has focussed on the softmax cross-entropy,
double-distillation may be useful for a broader class of losses,
e.g.,
\emph{order-weighted losses} as explored in~\citet{Usunier:2009,Reddi:2019}.

\section{Experimental results}
\label{sec:experiments}
We now present experiments illustrating three key points:
\begin{enumerate}[label=(\roman*),itemsep=0pt,topsep=0pt]
    \item we show that distilling with true (Bayes) class-probabilities improves generalisation over one-hot labels,
    validating
    our statistical view of distillation.

    \item we illustrate our bias-variance tradeoff on synthetic and real-world datasets,
    confirming that teachers with good estimates of $\pStar$ can be usefully distilled.

    \item we finally show that double-distillation performs well on real-world multiclass retrieval datasets,
    confirming the broader value in our statistical perspective.
\end{enumerate}

\begin{figure*}[!t]
    \centering
    
    \subfigure[Student AUC versus \# of training samples.]{
    \includegraphics[scale=0.4,valign=t]{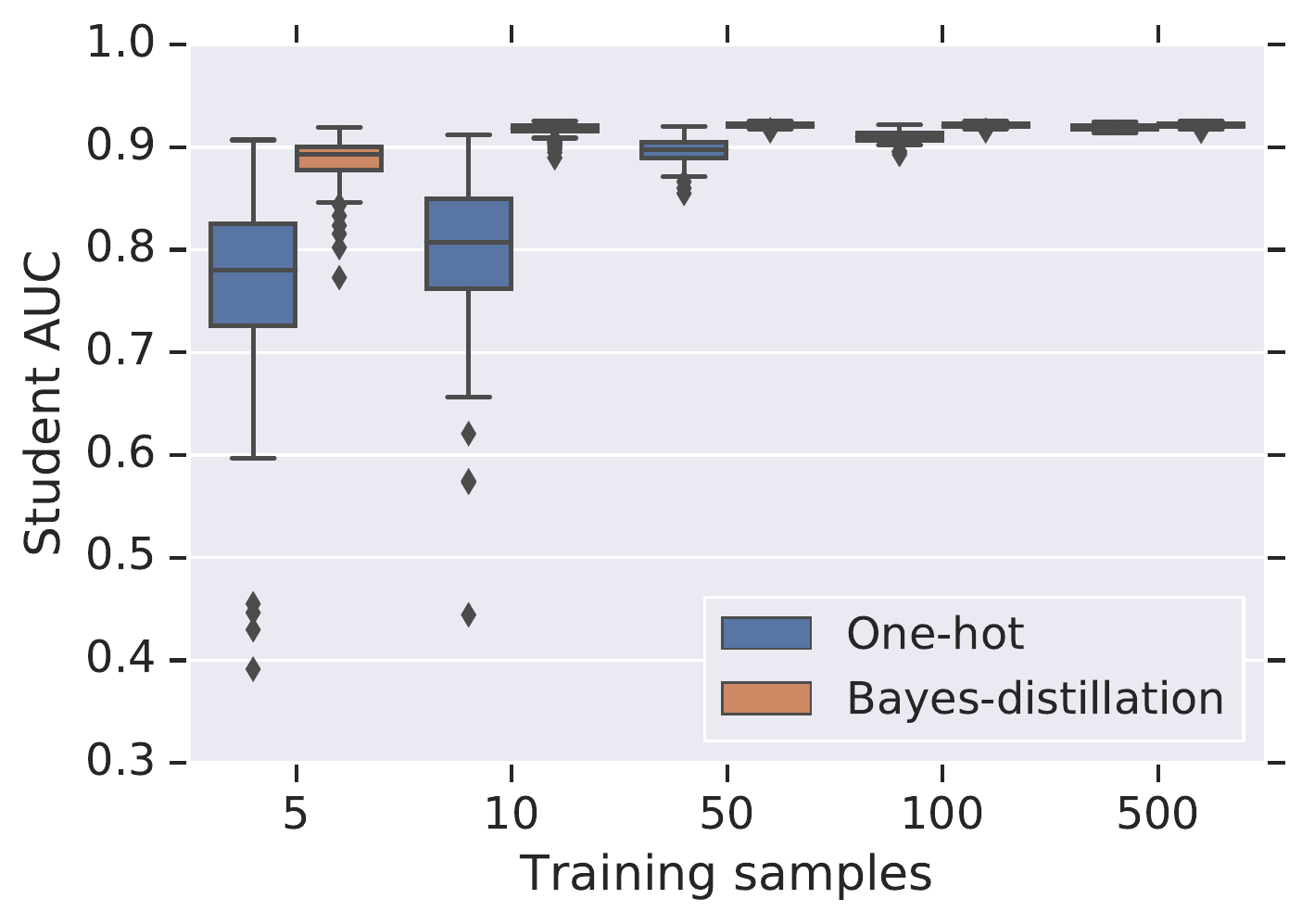}
    \label{fig:gaussian_distillation}
    }
%
    \qquad%
    \subfigure[Student AUC versus class separation.]{
    \includegraphics[scale=0.4,valign=t]{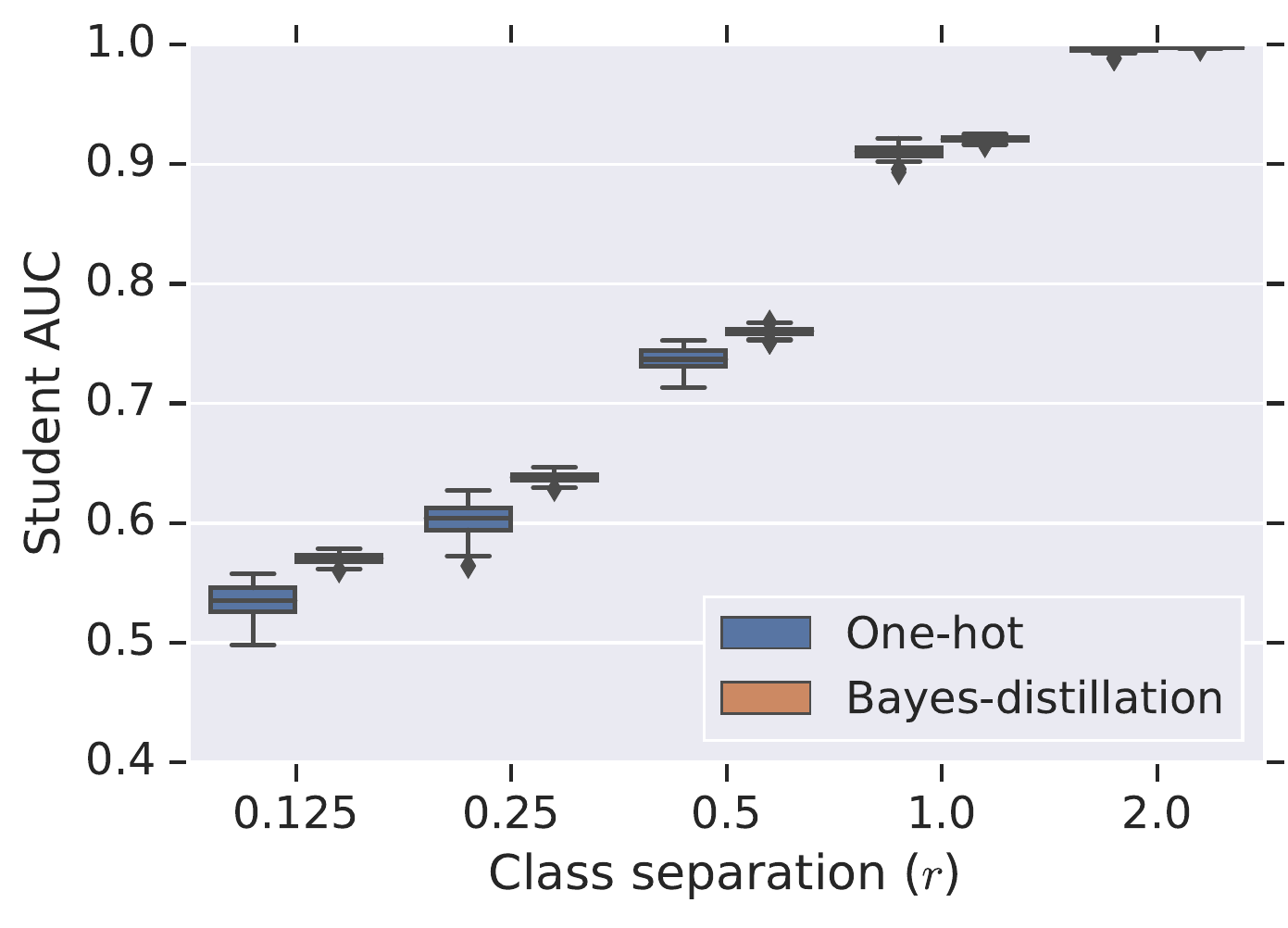}
    \label{fig:gaussian_distillation_vs_r}
    }
    
    \caption{Distillation with Bayes-teacher on synthetic data comprising Gaussian class-conditionals.
Distillation offers a noticeable gain over the standard one-hot encoding,
particularly in the small-sample regime (left),
and when the underlying problem is noisier (right).
    }
\end{figure*}

    


%
\subsection{Is Bayes a good teacher for distillation?}
\label{sec:experiments-bayes}

To illustrate our statistical perspective, we conduct a synthetic experiment where $\pStar( x )$ is known,
and show that distilling these Bayes class-probabilities benefits learning.

We generate training samples $S = \{ ( x_n, y_n ) \}_{i = 1}^N$ from a distribution $\Pr$ comprising class-conditionals which are each 10-dimensional Gaussians,
with means $\pm ( 1, 1, \ldots, 1 )$ respectively.
By construction, 
the Bayes class-probability distribution is 
$\pStar( x ) = ( \Pr( y = 0 \mid x ), \Pr( y = 1 \mid x ) )$
where
$\Pr( y = 1 \mid x ) = \sigma( (\theta^*)^{\tup}  x ) $,
for $\theta^* \defEq ( 2, 2, \ldots, 2 )$.

We compare two training procedures:
standard logistic regression on $S$, 
and \emph{Bayes-distilled} logistic regression using $\pStar( x_n )$ per~\eqref{eqn:bayes-distillation}.
Logistic regression is well-specified for this problem,
i.e.,
as $N \to \infty$,
the standard learner will learn $\theta^*$.
However, 
we will demonstrate
that on finite samples,
the Bayes-distilled learner's knowledge of $\pStar( x_n )$ will be beneficial.
We reiterate that while this learner could trivially memorise the training $\pStar( x_n )$,
this would not generalise.

Figure~\ref{fig:gaussian_distillation} compares the performance of these two approaches for varying training set sizes, where for each training set size we perform $100$ independent trials and measure the AUC-ROC on a test set of $10^4$ samples.
We observe two key trends:
first, Bayes-distillation generally offers a noticeable gain over the standard one-hot encoding,
in line with our theoretical guarantee of low variance.

Second, both methods see improved performance with more samples, but the gains are greater for the one-hot encoding.
This in line with our intuition that distillation effectively augments each training sample:
when $N$ is large to begin with, the marginal gain of such augmentation is minimal.

Figure~\ref{fig:gaussian_distillation_vs_r} continues the exploration of this setting.
We now vary the distance $r$ between the means of each of the Gaussians.
When $r$ is small, the two distributions grow closer together, making the classification problem more challenging.
We thus observe that both methods see worse performance as $r$ is smaller.
At the same time, smaller $r$ makes the one-hot labels have higher variance compared to the Bayes class-probabilities.
Consequently, the gains of distillation over the one-hot encoding are greater for this setting,
in line with our guarantee on the lower-variance Bayes-distilled risk aiding generalisation (Proposition~\ref{prop:svp}).

As a final experiment, we verify the claim that teacher accuracy does \emph{not} suffice for improving student generalisation,
since this does not necessarily correlate with the quality of the teacher's probability estimates.
We assess this by artificially distorting the teacher probabilities so as to perfectly preserve teacher accuracy,
while degrading their approximation to $\pStar$.
Appendix~\ref{sec:distorted_bayes_distillation} presents plots confirming that such degradation progressively reduces the gains of distillation.

\subsection{Illustration of bias-variance tradeoff}

We next illustrate our analysis 
on the bias-variance tradeoff in distillation
from~\S\ref{sec:bias-variance} on {synthetic and real-world} datasets.

\textbf{Synthetic}.
We now train a series of increasingly complex teacher models $\pTeach$,
and assess their resulting distillation benefit on a synthetic problem.
Here, the data is sampled from a marginal $\Pr( x )$ which is a zero-mean isotropic Gaussian in 2D.
The class-probability function is given by $\eta( x ) = \sigma( 2 \cdot (\| x \|_{\infty} - 0.5) )$, so that the negatives are concentrated in a rectangular slab.

We consider teachers that are 
random forests of
a fixed depth $d \in \{ 1, 2, 4, 8, 16 \}$, with $3$ base estimators.
Increasing $d$ has the effect of reducing teacher \emph{bias} (since the class of depth-$d$ trees can better approximate $\pStar$),
but increasing teacher \emph{variance} (since the class of depth-$d$ trees can induce complex decision boundaries).
For fixed $d$, we train a teacher model on the given training sample $S$ (with $N = 100$).
We then distill the teacher predictions to a student model, which is a depth $4$ tree.
For each such teacher, we compute its MSE,
as well as the test set AUC of the corresponding distilled student.
We repeat this for $100$ independent trials.

\begin{figure}[!t]
    \centering
    
    \subfigure[Teacher depth versus MSE.]{
    \includegraphics[scale=0.47]{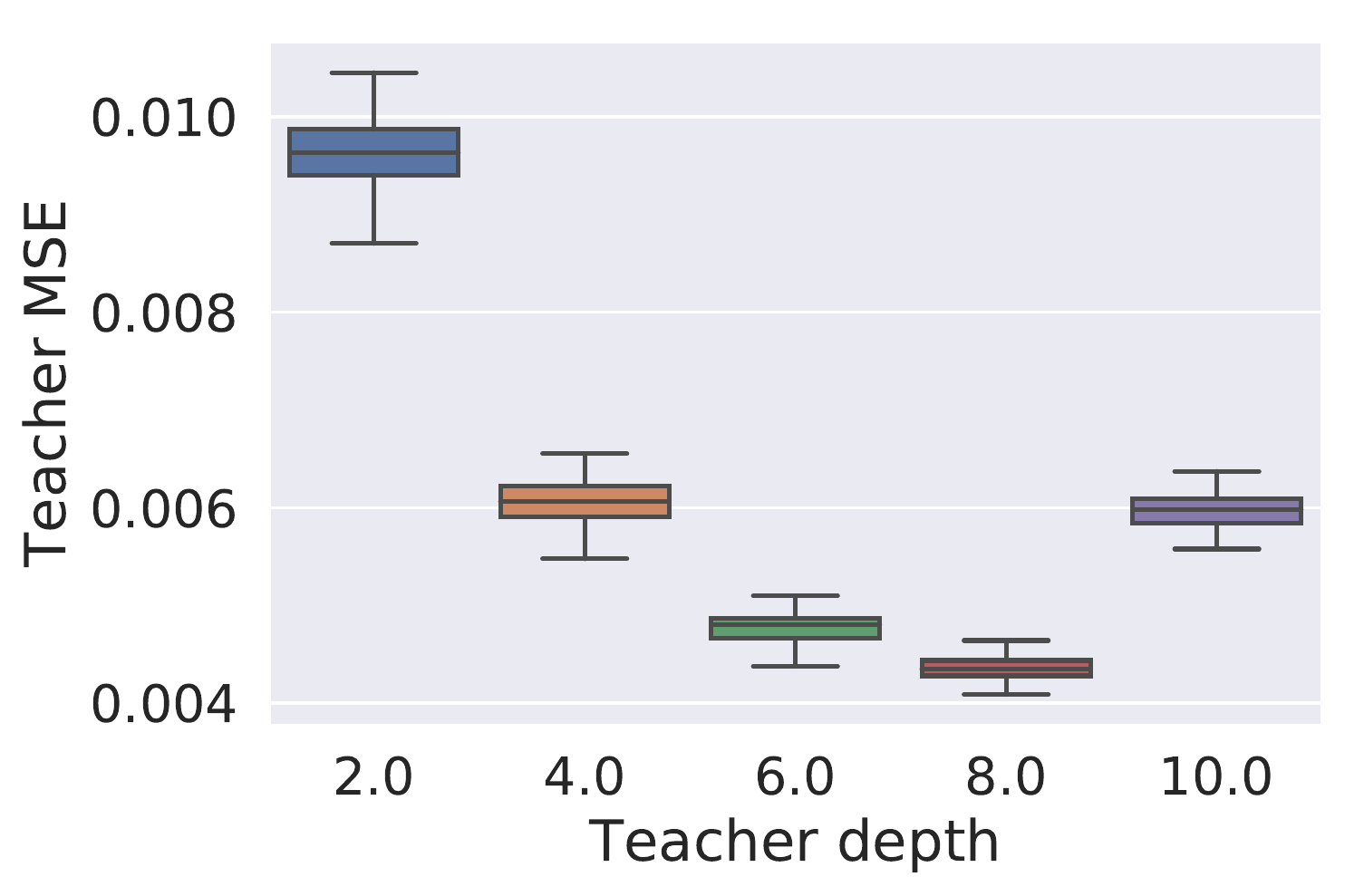}
    \label{fig:gaussian_distillation_tree_depth_vs_mse}
    }%
    \subfigure[Teacher depth versus student AUC.]{
    \includegraphics[scale=0.47]{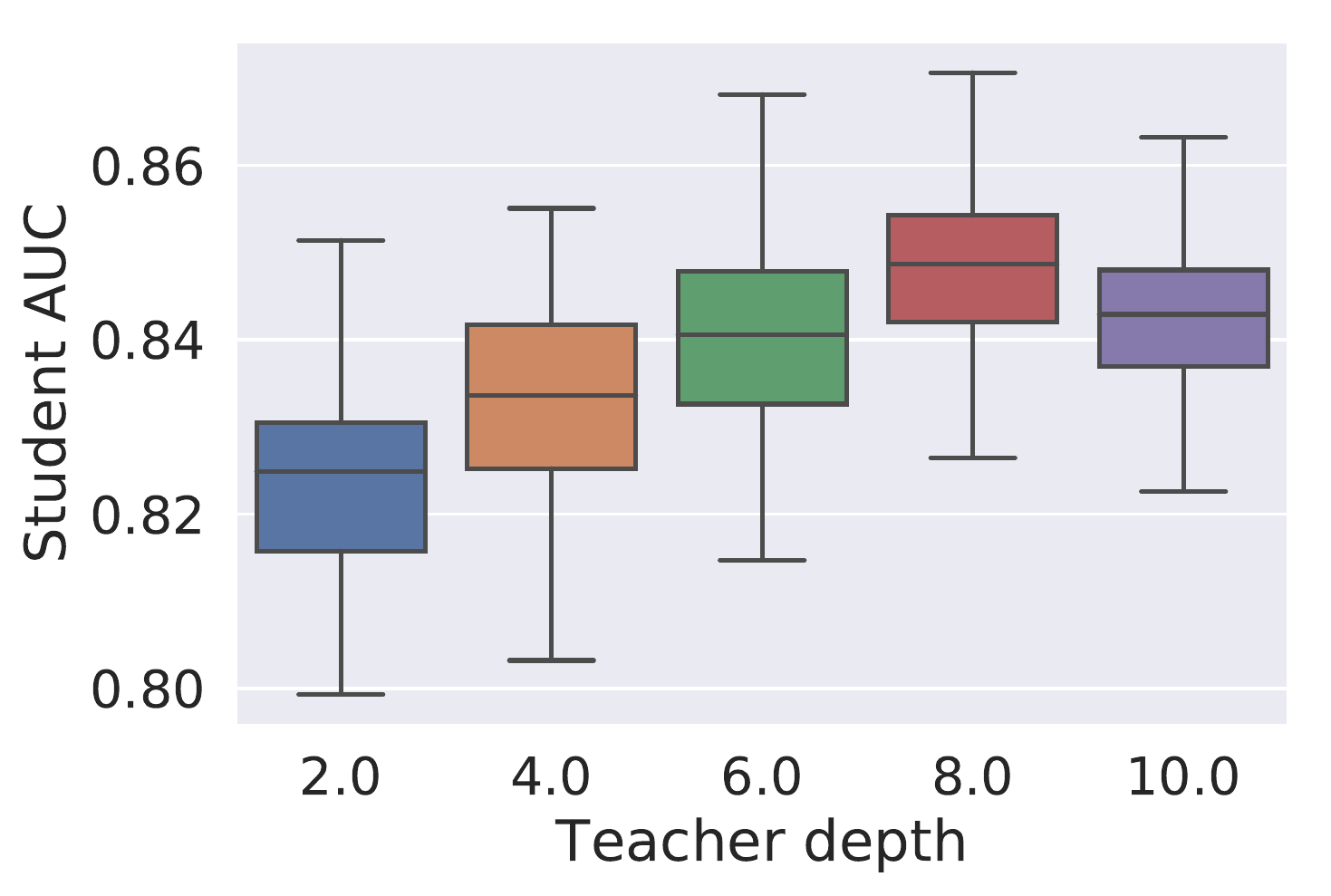}
    \label{fig:gaussian_distillation_tree_depth_vs_auc}
    }    

    \caption{Relationship between depth of teacher's decision tree model (model complexity) and its MSE in modelling $\pStar$, as well as the AUC of the resulting distilled student, on a synthetic 
    problem.
    There is an optimal depth $d = 8$ at which the teacher achieves the best MSE approximation of $\pStar$.
    In keeping with the theory, 
    this corresponds to the teacher whose resulting student generalises the best.}
\end{figure}

\begin{figure}[!t]
    \centering
    
    \subfigure[Synthetic dataset.]{
    \includegraphics[scale=0.47]{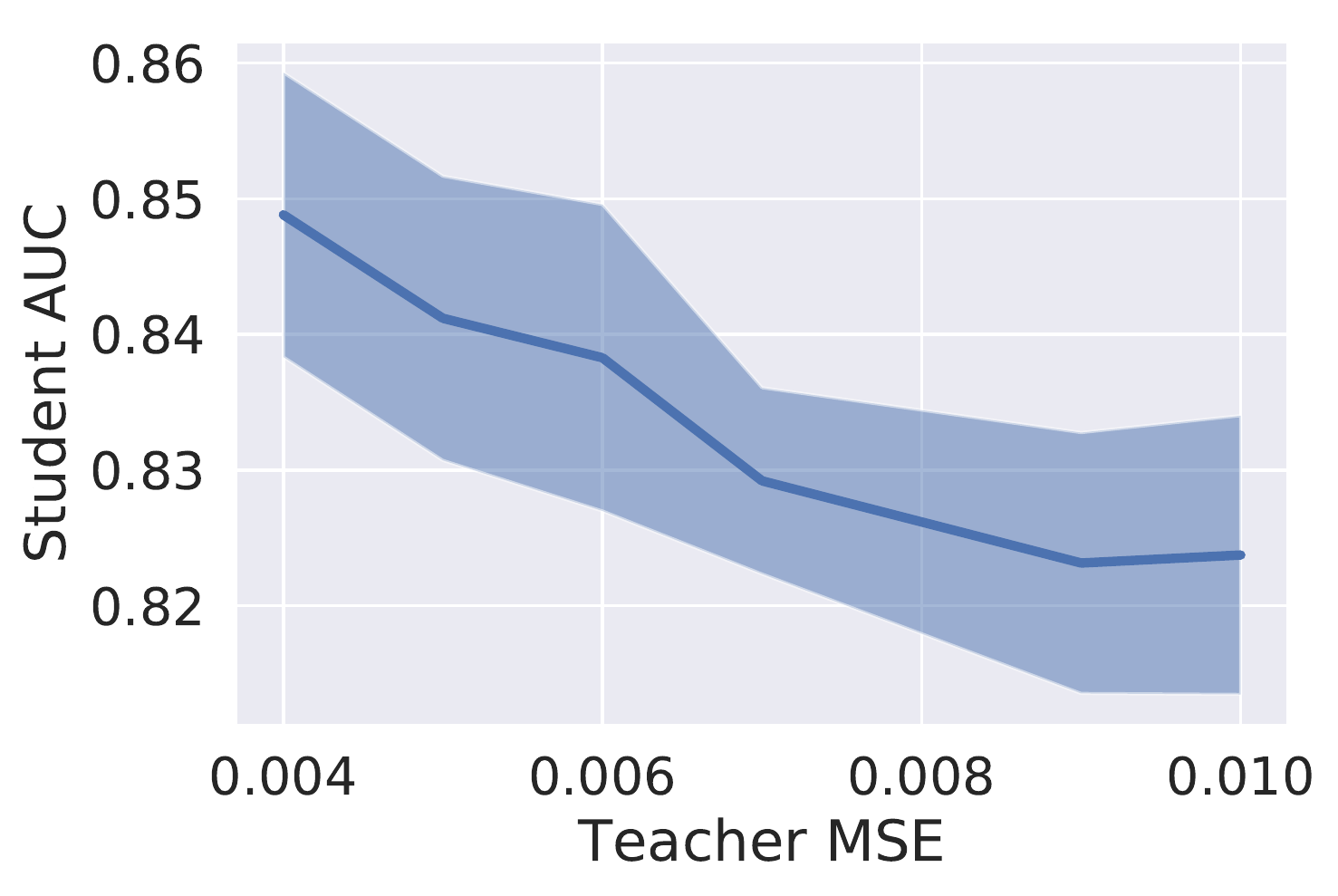}
    \label{fig:gaussian_distillation_tree_bias_variance}
    }%
    \subfigure[Fashion MNIST dataset.]{
    \includegraphics[scale=0.47]{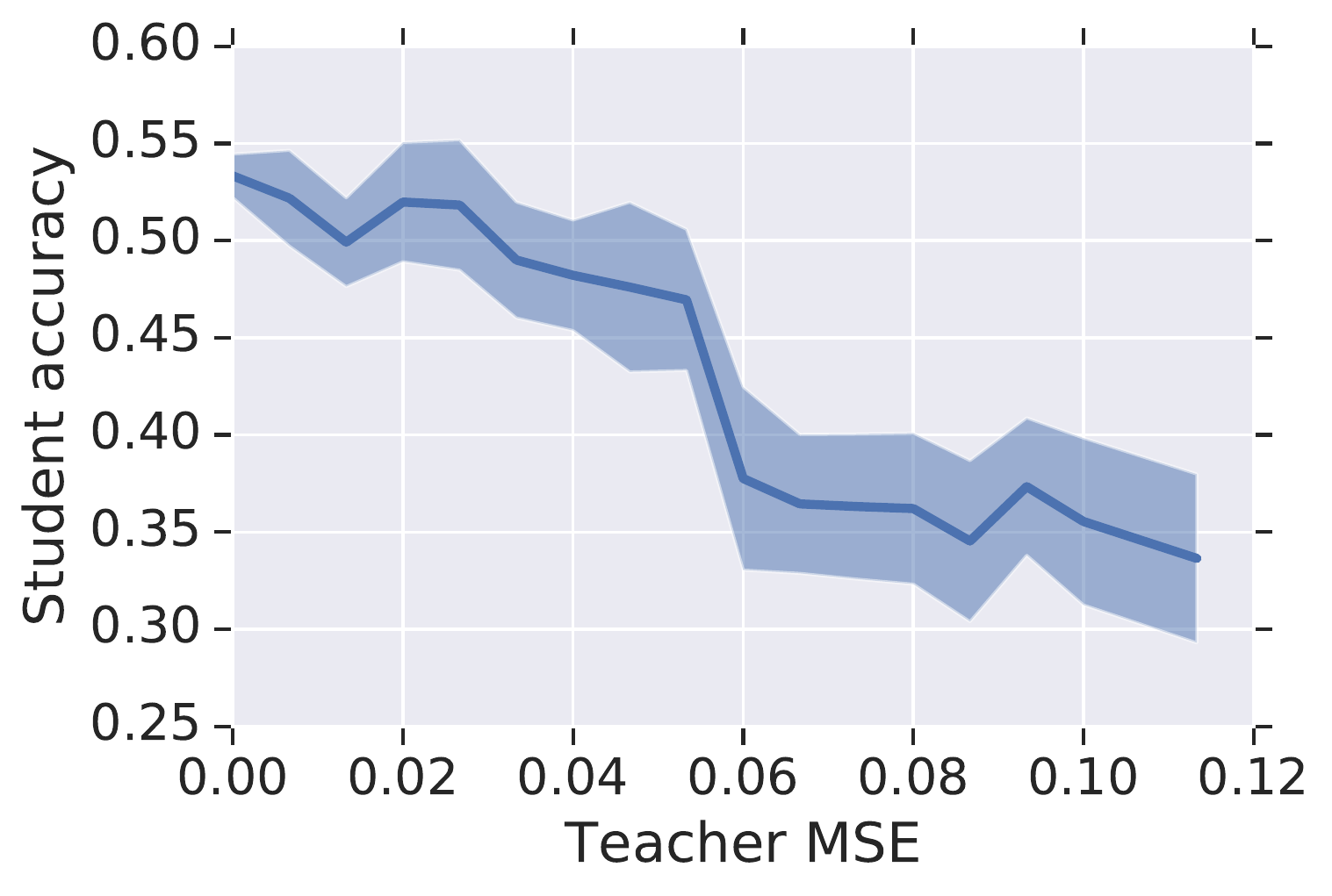}
    \label{fig:fmnist_distillation_bias_variance}
    }    

    \caption{Relationship between teacher's MSE against true class-probability and student's test set AUC.
    In keeping with the theory, teachers which better approximate $\pStar$ in an MSE sense
    yield better students.}
\end{figure}

Figure~\ref{fig:gaussian_distillation_tree_depth_vs_mse} and~\ref{fig:gaussian_distillation_tree_depth_vs_auc}
show how the teacher's depth affects its MSE in modelling $\pStar$, as well as the AUC of the resulting distilled student.
There is an optimal depth $d = 8$ at which the teacher achieves the best MSE approximation of $\pStar$.
In keeping with the theory, this also corresponds to the teacher whose resulting student generalises the best.
Figure~\ref{fig:gaussian_distillation_tree_bias_variance} combines these plots to explicitly show the relationship between the the teacher's MSE and the student's AUC.
In line with the theory, more accurate estimates of $\pStar$ result in better students.

Note that at depth $10$, the teacher model is expected to have lower bias;
however, it results in a slightly worse distilled student.
This verifies that \emph{one may favour a higher-bias teacher}
\emph{if it has lower variance}: a teacher may achieve a lower MSE -- and thus distill better -- by slightly increasing its bias while lowering variance. See Appendix~\ref{app:experiments-bias-variance} for additional bias-variance experiments on synthetic data.

\textbf{Fashion MNIST}.
It is challenging to assess the bias-variance tradeoff on real-world datasets,
where the Bayes $\pStar( x )$ is unknown.
As a proxy, 
we take the fashion MNIST dataset,
and treat a powerful teacher model as our $\pStar$.
We train an MLP teacher with two hidden layers with $128$ and $10$ dimensions. 
This achieves a test accuracy of $\sim 89\%$.

We then inject bias and noise per~\eqref{eqn:noisy-bayes},
and distill the result to a linear logistic regression model.
To amplify the effects of distillation, we constrain the student by only offering it the top $2500$ samples that the original teacher deems most uncertain.
Figures~\ref{fig:fmnist_distillation_bias_variance} 
demonstrates similar trend to the synthetic dataset,
with the best MSE approximator to the original teacher generally yielding the best student.

\textbf{CIFAR-100}.
\label{sec:bias-variance-cifar100}
We verify that accurate probabilty estimation by the teacher strongly influences student generalisation, and that this can be at odds with accuracy.
We revisit the plots introduced in Figure~\ref{fig:cifar100-distillation}.
Here, we train ResNets of varying depths on the CIFAR-100 dataset,
and use these as teachers to distill to a student ResNet of fixed depth $8$.
Figure~\ref{fig:cifar100_acc} reveals that the teacher model gets increasingly more accurate as its depth increases;
\emph{however}, the corresponding log-loss 
starts increasing beyond a depth of $14$.
This indicates the teacher's probability estimates become progressively poorer approximations of the Bayes class-probability distribution $\pStar( x )$. The accuracy of the student model also degrades beyond a teacher depth of $14$, reflecting the bias-variance bound in Proposition~\ref{prop:bias-variance}.

    


%

\subsection{Double-distillation for \XLR{}}
\label{sec:double-dist-expts}

Our final set of experiments confirm the value in our double-distillation objective in~\eqref{eqn:double-distillation}.
To do so, we use  the
{\sc AmazonCat-13K} and {\sc Amazon-670K}
 benchmark datasets for \XLR{}~\citep{McAuley:2013, Bhatia:2015}.
The data is multi\emph{label};
following~\citet{Reddi:2019}, we make it multiclass by
creating a single example $(x, y)$ for each label $y$ associated with $x$.

We construct a ``teacher'' model using a feedforward network with a single (linear) hidden layer of width $512$,
trained to minimise the softmax cross-entropy loss.
We then construct a ``student'' model using the same architecture, but with a hidden layer of width $8$ for {\sc AmazonCat-13K} and $64$ for {\sc Amazon-670K}
because  {\sc Amazon-670K} is significantly larger than {\sc AmazonCat-13K} (670k vs 13k labels).
This student model is compared to a distilled student,
where the teacher logits are used in place of the one-hot training labels.
Both methods are then compared to the double-distillation objective,
where the teacher logits are used to smooth the negatives in the softmax per~\eqref{eqn:generalised-xent} and~\eqref{eqn:double-distillation}.

We compare all methods using
the precision@$k$ metric
with $k \in \{ 1, 3, 5 \}$,
averaging these over multiple runs.
Table~\ref{tbl:double-distillation} summarises our findings.
We see that distillation offers a small but consistent bump in performance over the student baseline.
Double-distillation further improves upon this,
especially at the head of the prediction (P@1 and P@3),
confirming the value of weighing negatives differently.
The gains are particularly significant on {\sc AmazonCat-13K},
where the double-distilled student can \emph{improve} upon the teacher model itself.
Overall, our findings illustrate the broader value of the statistical perspective of distillation.

\begin{table*}[t!]
    \centering
    
    \begin{minipage}[t]{.475\linewidth}
    \begin{tabular}{@{}llll@{}}
        \toprule
        {\bf Method}                  & {\bf P@1} & {\bf P@3} & {\bf P@5} \\
        \toprule
        Teacher                       & 0.8495 & 0.7412 & 0.6109 \\
        \midrule
        Student                       & 0.7913 & 0.6156 & 0.4774 \\
        Student + distillation        & 0.8131 & 0.6363 & 0.4918 \\
        Student + double-distillation & \bf 0.8560 & \bf 0.7148 & \bf 0.5715 \\
        \bottomrule
    \end{tabular}
    
    \end{minipage}
    
    \qquad
    \vspace{\baselineskip}
    
    \begin{minipage}[t]{.475\linewidth}
    \begin{tabular}{@{}llll@{}}
        \toprule
        {\bf Method}                  & {\bf P@1} & {\bf P@3} & {\bf P@5} \\
        \toprule
        Teacher                       &  0.3983 &  0.3598 & 0.3298 \\
        \midrule
        Student                       &  0.3307 &  0.3004 & 0.2753 \\
        Student + distillation        &  0.3461 &  0.3151 & \bf 0.2892 \\
        Student + double-distillation &  \bf 0.3480 &  \bf 0.3161 & 0.2865 \\
        \bottomrule
    \end{tabular}
    
    \end{minipage}
    
    \caption{Precision@$k$ metrics for double-distillation objective against standard distillation and student baseline on {\sc AmazonCat-13K} (left) and {\sc AmazonCat-670K} (right).
    With double-distillation, the student is seen to significantly improve performance over training with one-hot labels, 
    but also a distilled model which applies uniform weighting of all negatives.
    }
    \label{tbl:double-distillation}
\end{table*}

\section{Conclusion}
We presented a statistical perspective on distillation,
building on a simple observation:
\emph{distilling the Bayes class-probabilities yields a more reliable estimate of the population risk}.
Viewing distillation in this light, we formalised a bias-variance tradeoff to quantify the effect of approximate teacher class-probability estimates on student generalisation,
and also studied a novel application of distillation to \XLR{}.
Towards developing a comprehensive understanding of distillation,
studying the optimisation aspects of this viewpoint,
and the setting of overparametrised teacher models~\citep{Zhang:2018} would be of interest.


%
\bibliography{references}

\begin{thebibliography}{59}
\providecommand{\natexlab}[1]{#1}
\providecommand{\url}[1]{\texttt{#1}}
\expandafter\ifx\csname urlstyle\endcsname\relax
  \providecommand{\doi}[1]{doi: #1}\else
  \providecommand{\doi}{doi: \begingroup \urlstyle{rm}\Url}\fi

\bibitem[Anil et~al.(2018)Anil, Pereyra, Passos, Ormandi, Dahl, and
  Hinton]{Anil:2018}
Rohan Anil, Gabriel Pereyra, Alexandre Passos, Robert Ormandi, George~E. Dahl,
  and Geoffrey~E. Hinton.
\newblock Large scale distributed neural network training through online
  distillation.
\newblock In \emph{International Conference on Learning Representations}, 2018.
\newblock URL \url{https://openreview.net/forum?id=rkr1UDeC-}.

\bibitem[Ba and Caruana(2014)]{Ba:2014}
Jimmy Ba and Rich Caruana.
\newblock Do deep nets really need to be deep?
\newblock In Z.~Ghahramani, M.~Welling, C.~Cortes, N.~D. Lawrence, and K.~Q.
  Weinberger, editors, \emph{Advances in Neural Information Processing Systems
  27}, pages 2654--2662. Curran Associates, Inc., 2014.

\bibitem[Bennett(1962)]{Bennett:1962}
George Bennett.
\newblock Probability inequalities for the sum of independent random variables.
\newblock \emph{Journal of the American Statistical Association}, 57\penalty0
  (297):\penalty0 33--45, 1962.

\bibitem[Bhatia et~al.(2015)Bhatia, Jain, Kar, Varma, and Jain]{Bhatia:2015}
Kush Bhatia, Himanshu Jain, Purushottam Kar, Manik Varma, and Prateek Jain.
\newblock Sparse local embeddings for extreme multi-label classification.
\newblock In \emph{Advances in Neural Information Processing Systems}, pages
  730--738, 2015.

\bibitem[Breiman and Shang(1996)]{Breiman:1996}
Leo Breiman and Nong Shang.
\newblock Born again trees.
\newblock
  \url{https://pdfs.semanticscholar.org/b6ba/5374ed8e09845996626bc62cc6d938e83fee.pdf},
  1996.

\bibitem[Bruch(2019)]{Bruch:2019b}
Sebastian Bruch.
\newblock An alternative cross entropy loss for learning-to-rank.
\newblock \emph{CoRR}, abs/1911.09798, 2019.

\bibitem[Bruch et~al.(2019)Bruch, Wang, Bendersky, and Najork]{Bruch:2019}
Sebastian Bruch, Xuanhui Wang, Michael Bendersky, and Marc Najork.
\newblock An analysis of the softmax cross entropy loss for learning-to-rank
  with binary relevance.
\newblock In \emph{Proceedings of the 2019 ACM SIGIR International Conference
  on Theory of Information Retrieval}, ICTIR ’19, page 75–78, New York, NY,
  USA, 2019. Association for Computing Machinery.
\newblock ISBN 9781450368810.

\bibitem[Bucil\v{a} et~al.(2006)Bucil\v{a}, Caruana, and
  Niculescu-Mizil]{Bucilua:2006}
Cristian Bucil\v{a}, Rich Caruana, and Alexandru Niculescu-Mizil.
\newblock Model compression.
\newblock In \emph{Proceedings of the 12th ACM SIGKDD International Conference
  on Knowledge Discovery and Data Mining}, KDD '06, pages 535--541, New York,
  NY, USA, 2006. ACM.

\bibitem[Buja et~al.(2005)Buja, Stuetzle, and Shen]{Buja:2005}
Andreas Buja, Werner Stuetzle, and Yi~Shen.
\newblock Loss functions for binary class probability estimation and
  classification: {S}tructure and applications.
\newblock Technical report, UPenn, 2005.

\bibitem[Cao et~al.(2019)Cao, Wei, Gaidon, Ar{\'{e}}chiga, and Ma]{Cao:2019}
Kaidi Cao, Colin Wei, Adrien Gaidon, Nikos Ar{\'{e}}chiga, and Tengyu Ma.
\newblock Learning imbalanced datasets with label-distribution-aware margin
  loss.
\newblock In \emph{Advances in Neural Information Processing Systems 32: Annual
  Conference on Neural Information Processing Systems 2019, NeurIPS 2019, 8-14
  December 2019, Vancouver, BC, Canada}, pages 1565--1576, 2019.

\bibitem[{Celik} et~al.(2017){Celik}, {Lopez-Paz}, and {McDaniel}]{Celik:2017}
Z.~B. {Celik}, D.~{Lopez-Paz}, and P.~{McDaniel}.
\newblock Patient-driven privacy control through generalized distillation.
\newblock In \emph{2017 IEEE Symposium on Privacy-Aware Computing (PAC)}, pages
  1--12, Aug 2017.

\bibitem[Cohen et~al.(1999)Cohen, Schapire, and Singer]{Cohen:1999}
William~W. Cohen, Robert~E. Schapire, and Yoram Singer.
\newblock Learning to order things.
\newblock \emph{J. Artif. Intell. Res.}, 10:\penalty0 243--270, 1999.

\bibitem[Crammer and Singer(2002)]{Crammer:2002}
Koby Crammer and Yoram Singer.
\newblock On the algorithmic implementation of multiclass kernel-based vector
  machines.
\newblock \emph{J. Mach. Learn. Res.}, 2:\penalty0 265--292, March 2002.
\newblock ISSN 1532-4435.

\bibitem[Craven and Shavlik(1995)]{Craven:1995}
Mark~W. Craven and Jude~W. Shavlik.
\newblock Extracting tree-structured representations of trained networks.
\newblock In \emph{Proceedings of the 8th International Conference on Neural
  Information Processing Systems}, NIPS'95, pages 24--30, Cambridge, MA, USA,
  1995. MIT Press.

\bibitem[Czarnecki et~al.(2017)Czarnecki, Osindero, Jaderberg, Swirszcz, and
  Pascanu]{Czarnecki:2017}
Wojciech~M. Czarnecki, Simon Osindero, Max Jaderberg, Grzegorz Swirszcz, and
  Razvan Pascanu.
\newblock Sobolev training for neural networks.
\newblock In I.~Guyon, U.~V. Luxburg, S.~Bengio, H.~Wallach, R.~Fergus,
  S.~Vishwanathan, and R.~Garnett, editors, \emph{Advances in Neural
  Information Processing Systems 30}, pages 4278--4287. Curran Associates,
  Inc., 2017.

\bibitem[Dekel et~al.(2003)Dekel, Manning, and Singer]{Dekel:2004}
Ofer Dekel, Christopher~D. Manning, and Yoram Singer.
\newblock Log-linear models for label ranking.
\newblock In \emph{Advances in Neural Information Processing Systems 16}, pages
  497--504, 2003.

\bibitem[Dong et~al.(2019)Dong, Hou, Lu, and Zhang]{Dong:2019}
Bin Dong, Jikai Hou, Yiping Lu, and Zhihua Zhang.
\newblock Distillation $\approx$ early stopping? harvesting dark knowledge
  utilizing anisotropic information retrieval for overparameterized neural
  network, 2019.

\bibitem[Foster et~al.(2019)Foster, Greenberg, Kale, Luo, Mohri, and
  Sridharan]{Foster:2019}
Dylan~J Foster, Spencer Greenberg, Satyen Kale, Haipeng Luo, Mehryar Mohri, and
  Karthik Sridharan.
\newblock Hypothesis set stability and generalization.
\newblock In \emph{Advances in Neural Information Processing Systems 32}, pages
  6726--6736. Curran Associates, Inc., 2019.

\bibitem[Furlanello et~al.(2018)Furlanello, Lipton, Tschannen, Itti, and
  Anandkumar]{Furlanello:2018}
Tommaso Furlanello, Zachary~Chase Lipton, Michael Tschannen, Laurent Itti, and
  Anima Anandkumar.
\newblock Born-again neural networks.
\newblock In \emph{Proceedings of the 35th International Conference on Machine
  Learning, {ICML} 2018, Stockholmsm{\"{a}}ssan, Stockholm, Sweden, July 10-15,
  2018}, pages 1602--1611, 2018.

\bibitem[F{\"u}rnkranz et~al.(2008)F{\"u}rnkranz, H{\"u}llermeier,
  Loza~Menc{\'\i}a, and Brinker]{Furnkranz:2008}
Johannes F{\"u}rnkranz, Eyke H{\"u}llermeier, Eneldo Loza~Menc{\'\i}a, and
  Klaus Brinker.
\newblock Multilabel classification via calibrated label ranking.
\newblock \emph{Machine Learning}, 73\penalty0 (2):\penalty0 133--153, 2008.

\bibitem[Gotmare et~al.(2019)Gotmare, Keskar, Xiong, and Socher]{Gotmare:2019}
Akhilesh Gotmare, Nitish~Shirish Keskar, Caiming Xiong, and Richard Socher.
\newblock A closer look at deep learning heuristics: Learning rate restarts,
  warmup and distillation.
\newblock In \emph{International Conference on Learning Representations}, 2019.

\bibitem[Guo et~al.(2017)Guo, Pleiss, Sun, and Weinberger]{Guo:2017}
Chuan Guo, Geoff Pleiss, Yu~Sun, and Kilian~Q. Weinberger.
\newblock On calibration of modern neural networks.
\newblock In \emph{Proceedings of the 34th International Conference on Machine
  Learning, {ICML} 2017, Sydney, NSW, Australia, 6-11 August 2017}, pages
  1321--1330, 2017.

\bibitem[Hinton et~al.(2015)Hinton, Vinyals, and Dean]{Hinton:2015}
Geoffrey~E. Hinton, Oriol Vinyals, and Jeffrey Dean.
\newblock Distilling the knowledge in a neural network.
\newblock \emph{CoRR}, abs/1503.02531, 2015.

\bibitem[Jain et~al.(2019)Jain, Balasubramanian, Chunduri, and
  Varma]{Jain:2019}
Himanshu Jain, Venkatesh Balasubramanian, Bhanu Chunduri, and Manik Varma.
\newblock Slice: Scalable linear extreme classifiers trained on 100 million
  labels for related searches.
\newblock In \emph{Proceedings of the Twelfth ACM International Conference on
  Web Search and Data Mining}, WSDM '19, pages 528--536, New York, NY, USA,
  2019. ACM.

\bibitem[{Khan} et~al.(2019){Khan}, {Hayat}, {Zamir}, {Shen}, and
  {Shao}]{Khan:2019}
S.~{Khan}, M.~{Hayat}, S.~W. {Zamir}, J.~{Shen}, and L.~{Shao}.
\newblock Striking the right balance with uncertainty.
\newblock In \emph{2019 IEEE/CVF Conference on Computer Vision and Pattern
  Recognition (CVPR)}, pages 103--112, 2019.

\bibitem[{Lapin} et~al.(2018){Lapin}, {Hein}, and {Schiele}]{Lapin:2018}
M.~{Lapin}, M.~{Hein}, and B.~{Schiele}.
\newblock Analysis and optimization of loss functions for multiclass, top-k,
  and multilabel classification.
\newblock \emph{IEEE Transactions on Pattern Analysis and Machine
  Intelligence}, 40\penalty0 (7):\penalty0 1533--1554, July 2018.

\bibitem[{Li} and {Hoiem}(2018)]{Li:2018}
Z.~{Li} and D.~{Hoiem}.
\newblock Learning without forgetting.
\newblock \emph{IEEE Transactions on Pattern Analysis and Machine
  Intelligence}, 40\penalty0 (12):\penalty0 2935--2947, Dec 2018.
\newblock ISSN 1939-3539.

\bibitem[Li et~al.(2019)Li, Kamnitsas, and Glocker]{Li:2019}
Zeju Li, Konstantinos Kamnitsas, and Ben Glocker.
\newblock Overfitting of neural nets under class imbalance: Analysis and
  improvements for segmentation.
\newblock In Dinggang Shen, Tianming Liu, Terry~M. Peters, Lawrence~H. Staib,
  Caroline Essert, Sean Zhou, Pew-Thian Yap, and Ali Khan, editors,
  \emph{Medical Image Computing and Computer Assisted Intervention -- MICCAI
  2019}, pages 402--410, Cham, 2019. Springer International Publishing.

\bibitem[Liu et~al.(2016)Liu, Wen, Yu, and Yang]{Liu:2016}
Weiyang Liu, Yandong Wen, Zhiding Yu, and Meng Yang.
\newblock Large-margin softmax loss for convolutional neural networks.
\newblock In \emph{Proceedings of the 33rd International Conference on
  International Conference on Machine Learning - Volume 48}, ICML’16, page
  507–516. JMLR.org, 2016.

\bibitem[Liu et~al.(2017)Liu, Wen, Yu, Li, Raj, and Song]{Liu:2017}
Weiyang Liu, Yandong Wen, Zhiding Yu, Ming Li, Bhiksha Raj, and Le~Song.
\newblock Sphereface: Deep hypersphere embedding for face recognition.
\newblock In \emph{2017 {IEEE} Conference on Computer Vision and Pattern
  Recognition, {CVPR} 2017, Honolulu, HI, USA, July 21-26, 2017}, pages
  6738--6746, 2017.

\bibitem[Liu et~al.(2019)Liu, Jia, Tan, Vemulapalli, Zhu, Green, and
  Wang]{Liu:2019}
Yu~Liu, Xuhui Jia, Mingxing Tan, Raviteja Vemulapalli, Yukun Zhu, Bradley
  Green, and Xiaogang Wang.
\newblock Search to distill: Pearls are everywhere but not the eyes, 2019.

\bibitem[Lopez-Paz et~al.(2016)Lopez-Paz, Sch{\"o}lkopf, Bottou, and
  Vapnik]{Lopez-Paz:2016}
D.~Lopez-Paz, B.~Sch{\"o}lkopf, L.~Bottou, and V.~Vapnik.
\newblock Unifying distillation and privileged information.
\newblock In \emph{International Conference on Learning Representations
  (ICLR)}, November 2016.

\bibitem[Maurer and Pontil(2009)]{Maurer:2009}
Andreas Maurer and Massimiliano Pontil.
\newblock Empirical bernstein bounds and sample-variance penalization.
\newblock In \emph{{COLT} 2009 - The 22nd Conference on Learning Theory,
  Montreal, Quebec, Canada, June 18-21, 2009}, 2009.

\bibitem[McAuley and Leskovec(2013)]{McAuley:2013}
Julian McAuley and Jure Leskovec.
\newblock Hidden factors and hidden topics: Understanding rating dimensions
  with review text.
\newblock In \emph{Proceedings of the 7th ACM Conference on Recommender
  Systems}, RecSys ’13, page 165–172, New York, NY, USA, 2013. Association
  for Computing Machinery.
\newblock ISBN 9781450324090.

\bibitem[Mobahi et~al.(2020)Mobahi, Farajtabar, and Bartlett]{Mobahi:2020}
Hossein Mobahi, Mehrdad Farajtabar, and Peter~L. Bartlett.
\newblock Self-distillation amplifies regularization in hilbert space, 2020.

\bibitem[M{\"{u}}ller et~al.(2019)M{\"{u}}ller, Kornblith, and
  Hinton]{Muller:2019}
Rafael M{\"{u}}ller, Simon Kornblith, and Geoffrey~E. Hinton.
\newblock When does label smoothing help?
\newblock In \emph{Advances in Neural Information Processing Systems 32: Annual
  Conference on Neural Information Processing Systems 2019, NeurIPS 2019, 8-14
  December 2019, Vancouver, BC, Canada}, pages 4696--4705, 2019.

\bibitem[Nayak et~al.(2019)Nayak, Mopuri, Shaj, Radhakrishnan, and
  Chakraborty]{Nayak:2019}
Gaurav~Kumar Nayak, Konda~Reddy Mopuri, Vaisakh Shaj, Venkatesh~Babu
  Radhakrishnan, and Anirban Chakraborty.
\newblock Zero-shot knowledge distillation in deep networks.
\newblock In \emph{Proceedings of the 36th International Conference on Machine
  Learning, {ICML} 2019, 9-15 June 2019, Long Beach, California, {USA}}, pages
  4743--4751, 2019.

\bibitem[Papernot et~al.(2016)Papernot, McDaniel, Wu, Jha, and
  Swami]{Papernot:2016}
Nicolas Papernot, Patrick~D. McDaniel, Xi~Wu, Somesh Jha, and Ananthram Swami.
\newblock Distillation as a defense to adversarial perturbations against deep
  neural networks.
\newblock In \emph{{IEEE} Symposium on Security and Privacy, {SP} 2016, San
  Jose, CA, USA, May 22-26, 2016}, pages 582--597, 2016.

\bibitem[Phuong and Lampert(2019)]{Phuong:2019}
Mary Phuong and Christoph Lampert.
\newblock Towards understanding knowledge distillation.
\newblock In Kamalika Chaudhuri and Ruslan Salakhutdinov, editors,
  \emph{Proceedings of the 36th International Conference on Machine Learning},
  volume~97 of \emph{Proceedings of Machine Learning Research}, pages
  5142--5151, Long Beach, California, USA, 09--15 Jun 2019. PMLR.

\bibitem[Radosavovic et~al.(2018)Radosavovic, Doll{\'{a}}r, Girshick, Gkioxari,
  and He]{Radosavovic:2018}
Ilija Radosavovic, Piotr Doll{\'{a}}r, Ross~B. Girshick, Georgia Gkioxari, and
  Kaiming He.
\newblock Data distillation: Towards omni-supervised learning.
\newblock In \emph{2018 {IEEE} Conference on Computer Vision and Pattern
  Recognition, {CVPR} 2018, Salt Lake City, UT, USA, June 18-22, 2018}, pages
  4119--4128, 2018.

\bibitem[Reddi et~al.(2019)Reddi, Kale, Yu, Holtmann-Rice, Chen, and
  Kumar]{Reddi:2019}
Sashank~J. Reddi, Satyen Kale, Felix Yu, Daniel Holtmann-Rice, Jiecao Chen, and
  Sanjiv Kumar.
\newblock Stochastic negative mining for learning with large output spaces.
\newblock In \emph{Proceedings of Machine Learning Research}, volume~89 of
  \emph{Proceedings of Machine Learning Research}, pages 1940--1949. PMLR,
  16--18 Apr 2019.

\bibitem[Rothfuss et~al.(2019)Rothfuss, Ferreira, Boehm, Walther, Ulrich,
  Asfour, and Krause]{Rothfuss:2019}
Jonas Rothfuss, Fabio Ferreira, Simon Boehm, Simon Walther, Maxim Ulrich, Tamim
  Asfour, and Andreas Krause.
\newblock Noise regularization for conditional density estimation, 2019.

\bibitem[Rudin(2009)]{Rudin:2009}
Cynthia Rudin.
\newblock The {P}-{N}orm {P}ush: A simple convex ranking algorithm that
  concentrates at the top of the list.
\newblock \emph{Journal of Machine Learning Research}, 10:\penalty0 2233--2271,
  Oct 2009.

\bibitem[Rusu et~al.(2016)Rusu, Colmenarejo, G{\"{u}}l{\c{c}}ehre, Desjardins,
  Kirkpatrick, Pascanu, Mnih, Kavukcuoglu, and Hadsell]{Rusu:2016}
Andrei~A. Rusu, Sergio~Gomez Colmenarejo, {\c{C}}aglar G{\"{u}}l{\c{c}}ehre,
  Guillaume Desjardins, James Kirkpatrick, Razvan Pascanu, Volodymyr Mnih,
  Koray Kavukcuoglu, and Raia Hadsell.
\newblock Policy distillation.
\newblock In \emph{4th International Conference on Learning Representations,
  {ICLR} 2016, San Juan, Puerto Rico, May 2-4, 2016, Conference Track
  Proceedings}, 2016.

\bibitem[Savage(1971)]{Savage:1971}
Leonard~J. Savage.
\newblock Elicitation of personal probabilities and expectations.
\newblock \emph{Journal of the American Statistical Association}, 66\penalty0
  (336):\penalty0 783--801, 1971.
\newblock ISSN 01621459.

\bibitem[Schervish(1989)]{Schervish:1989}
Mark~J. Schervish.
\newblock A general method for comparing probability assessors.
\newblock \emph{Ann. Statist.}, 17\penalty0 (4):\penalty0 1856--1879, 12 1989.

\bibitem[Szegedy et~al.(2016)Szegedy, Vanhoucke, Ioffe, Shlens, and
  Wojna]{Szegedy:2016}
Christian Szegedy, Vincent Vanhoucke, Sergey Ioffe, Jonathon Shlens, and
  Zbigniew Wojna.
\newblock Rethinking the inception architecture for computer vision.
\newblock In \emph{2016 {IEEE} Conference on Computer Vision and Pattern
  Recognition, {CVPR} 2016, Las Vegas, NV, USA, June 27-30, 2016}, pages
  2818--2826, 2016.

\bibitem[Tang and Wang(2018)]{Tang:2018}
Jiaxi Tang and Ke~Wang.
\newblock Ranking distillation: Learning compact ranking models with high
  performance for recommender system.
\newblock In \emph{Proceedings of the 24th ACM SIGKDD International Conference
  on Knowledge Discovery \& Data Mining}, KDD ’18, page 2289–2298, New
  York, NY, USA, 2018. Association for Computing Machinery.
\newblock ISBN 9781450355520.

\bibitem[Tang et~al.(2020)Tang, Shivanna, Zhao, Lin, Singh, Chi, and
  Jain]{Tang:2020}
Jiaxi Tang, Rakesh Shivanna, Zhe Zhao, Dong Lin, Anima Singh, Ed~H. Chi, and
  Sagar Jain.
\newblock Understanding and improving knowledge distillation.
\newblock \emph{CoRR}, abs/2002.03532, 2020.
\newblock URL \url{https://arxiv.org/abs/2002.03532}.

\bibitem[Tang et~al.(2016)Tang, Wang, and Zhang]{Tang:2016}
Zhiyuan Tang, Dong Wang, and Zhiyong Zhang.
\newblock Recurrent neural network training with dark knowledge transfer.
\newblock In \emph{2016 {IEEE} International Conference on Acoustics, Speech
  and Signal Processing, {ICASSP} 2016, Shanghai, China, March 20-25, 2016},
  pages 5900--5904, 2016.

\bibitem[Usunier et~al.(2009)Usunier, Buffoni, and Gallinari]{Usunier:2009}
Nicolas Usunier, David Buffoni, and Patrick Gallinari.
\newblock Ranking with ordered weighted pairwise classification.
\newblock In \emph{Proceedings of the 26th Annual International Conference on
  Machine Learning}, ICML '09, pages 1057--1064, New York, NY, USA, 2009. ACM.

\bibitem[Vembu and G{\"a}rtner(2011)]{Vembu:2011}
Shankar Vembu and Thomas G{\"a}rtner.
\newblock \emph{Label Ranking Algorithms: A Survey}, pages 45--64.
\newblock Springer Berlin Heidelberg, Berlin, Heidelberg, 2011.

\bibitem[{Wang} et~al.(2018){Wang}, {Cheng}, {Liu}, and {Liu}]{Wang:2018}
F.~{Wang}, J.~{Cheng}, W.~{Liu}, and H.~{Liu}.
\newblock Additive margin softmax for face verification.
\newblock \emph{IEEE Signal Processing Letters}, 25\penalty0 (7):\penalty0
  926--930, 2018.

\bibitem[Xie et~al.(2019)Xie, Hovy, Luong, and Le]{Xie:2019}
Qizhe Xie, Eduard Hovy, Minh-Thang Luong, and Quoc~V. Le.
\newblock Self-training with noisy student improves imagenet classification,
  2019.

\bibitem[Xue et~al.(2013)Xue, Li, and Gong]{Xue:2013}
Jian Xue, Jinyu Li, and Yifan Gong.
\newblock Restructuring of deep neural network acoustic models with singular
  value decomposition.
\newblock In \emph{Interspeech}, January 2013.

\bibitem[Yang et~al.(2019)Yang, Xie, Qiao, and Yuille]{Yang:2019}
Chenglin Yang, Lingxi Xie, Siyuan Qiao, and Alan~L. Yuille.
\newblock Training deep neural networks in generations: {A} more tolerant
  teacher educates better students.
\newblock In \emph{The Thirty-Third {AAAI} Conference on Artificial
  Intelligence, {AAAI} 2019}, pages 5628--5635, 2019.

\bibitem[{Yim} et~al.(2017){Yim}, {Joo}, {Bae}, and {Kim}]{Yim:2017}
J.~{Yim}, D.~{Joo}, J.~{Bae}, and J.~{Kim}.
\newblock A gift from knowledge distillation: Fast optimization, network
  minimization and transfer learning.
\newblock In \emph{2017 IEEE Conference on Computer Vision and Pattern
  Recognition (CVPR)}, pages 7130--7138, July 2017.

\bibitem[Yun et~al.(2014)Yun, Raman, and Vishwanathan]{Yun:2014}
Hyokun Yun, Parameswaran Raman, and S.~V.~N. Vishwanathan.
\newblock Ranking via robust binary classification.
\newblock In \emph{Advances in Neural Information Processing Systems 27: Annual
  Conference on Neural Information Processing Systems 2014, December 8-13 2014,
  Montreal, Quebec, Canada}, pages 2582--2590, 2014.

\bibitem[Zhang et~al.(2018)Zhang, Cisse, Dauphin, and Lopez-Paz]{Zhang:2018}
Hongyi Zhang, Moustapha Cisse, Yann~N. Dauphin, and David Lopez-Paz.
\newblock mixup: Beyond empirical risk minimization.
\newblock In \emph{International Conference on Learning Representations}, 2018.

\end{thebibliography}
\bibliographystyle{plainnat}

\clearpage
\appendix
\onecolumn

\begin{center}
  {\Large\bf Supplementary material for ``Why distillation helps: a statistical perspective''}
\end{center}

\section{Theory: additional results}
\label{sec:appendix-theory}

\begin{proposition}
\label{prop:mvue}
Suppose we have a teacher model $\pTeach$ with corresponding distilled empirical risk~\eqref{eqn:distillation-risk}. Furthermore, assume $\pTeach$ is unbiased, i.e., $\E{}{\pTeach(x)} = \pStar(x)$ for all $x \in \XCal$. Then, for any predictor $f \colon \XCal \to \Real^L$,
\begin{align*}
\E{}{ (\RDistBayes( f; S ) - R( f ) )^2 } &\leq \E{}{ ( \RDist( f; S ) - R( f ) )^2 } 
\end{align*}
for some constant $C > 0$.
\end{proposition}

\begin{proof}[Proof of Proposition~\ref{prop:mvue}]
Let $\hat{\Delta} \defEq \RDist( f; S ) - \RDistBayes( f ; S)$ and $\hat{\Delta}_* \defEq \RDistBayes( f ; S) - R(f)$.  
Then,
\begin{align*}
    \E{}{( \RDist( f; S ) - R( f ) )^2 } &= \E{}{ (\hat{\Delta} + \hat{\Delta}_*)^2 } .
\end{align*}
Note that $\E{\pTeach}{\hat{\Delta}} = 0$ since $\pTeach$ is an unbiased estimator of $\pStar$. Using this fact, we obtain the desired result as follows:
\begin{align*}
    \E{}{( \RDist( f; S ) - R( f ) )^2 } &= \E{}{ \hat{\Delta}^2 + \hat{\Delta}_*^2 } \\
    &\geq \E{}{\hat{\Delta}_*^2}.
\end{align*}
\end{proof}

\begin{proposition}
\label{prop:svp-phat}
Pick any bounded loss $\ell$.
Fix a hypothesis class $\FCal$ of predictors 
$f \colon \XCal \to \Real^L$,
with induced class $\HCal \subset [ 0, 1 ]^{\XCal}$ 
of functions $h( x ) \defEq \pTeach( x )^{\tup} \ell( f( x ) )$.
Suppose $\HCal$
has uniform covering number 
$\mathscr{N}_{\infty}$.
Then,
for any $\delta \in (0, 1)$,
with probability at least $1 - \delta$ over $S \sim \Pr^N$,
\begin{align*}
R( f )  &\leq  \RDist( f; S ) + \mathscr{O}\left( \sqrt{\tilde{\mathbb{V}}_N( f ) \cdot \frac{\log \frac{\mathscr{M}_N}{\delta}}{N}} + \frac{\log \frac{\mathscr{M}_N}{\delta}}{N} \right) + \mathscr{O}\left(\mathbb{E}{}{ \| \pTeach( \X ) - \pStar( \X ) \|_2}\right), \end{align*}
where $\mathscr{M}_N \defEq \mathscr{N}_{\infty}( \frac{1}{N}, \HCal, 2 N )$
and $\tilde{\mathbb{V}}_N( f )$ is the empirical variance of the loss values.
\end{proposition}

\begin{proof}[Proof of Proposition~\ref{prop:svp-phat}]
Let $\tilde{R}(f) = \E{}{\RDist( f; S )}$ and $\Delta \defEq \RDist( f; S ) - R( f )$. We note that with probability $1 - \delta$,
\begin{align}
\label{eq:prop5-step1}
\tilde{R}( f ) \leq \RDist( f; S ) + \mathscr{O}\left( \sqrt{ \tilde{\mathbb{V}}_N( f ) \cdot \frac{\log \frac{\mathscr{M}_N}{\delta}}{N}} + \frac{\log \frac{\mathscr{M}_N}{\delta}}{N} \right),
\end{align}
where $\mathscr{M}_N \defEq \mathscr{N}_{\infty}( \frac{1}{N}, \HCal, 2 N )$
and $\tilde{\mathbb{V}}_N( f )$ is the empirical variance of the loss values. Furthermore, the following holds
\begin{align*}
|\tilde{R}(f) - R(f)| &= \left|\E{}{\RDist( f; S )} - \E{}{\RDistBayes( f; S )}\right| \\
&\leq \E{}{ \| \pTeach( \X ) - \pStar( \X ) \|_2 \cdot \| \ell( f( \X ) ) \|_2}.
\end{align*}
Thus, we have
\begin{align}
\label{eq:prop5-step2}
R(f) \leq \tilde{R}(f) + C \cdot \E{}{ \| \pTeach( \X ) - \pStar( \X ) \|_2}.
\end{align}
for some constant $C > 0$. The result follows by combining \eqref{eq:prop5-step1} and \eqref{eq:prop5-step2}.
\end{proof}

\clearpage

\section{Experiments: additional results}

\subsection{Bias-variance tradeoff}
\label{app:experiments-bias-variance}

Continuing the same synthetic Gaussian data as in~\S\ref{sec:experiments-bayes},
we now consider a family of teachers of the form
\begin{equation}
    \label{eqn:noisy-bayes}
    \pTeach( x ) = (1 - \alpha) \cdot \Psi( (\theta^*)^{\tup} x + \sigma^2 \cdot \epsilon ) + \frac{\alpha}{2},
\end{equation}
where 
$\Psi( z ) \defEq (1 + e^{-z})^{-1}$ is the sigmoid,
$\alpha \in [ 0, 1 ]$,
$\sigma > 0$, and
and $\epsilon \sim \mathscr{N}( 0, 1 )$ comprises independent Gaussian noise.
Increasing $\alpha$
induces a \emph{bias} in the teacher's estimate of $\pStar( x )$,
while increasing $\sigma$ induces a \emph{variance} in the teacher over fresh draws.
Combined, these induce the teacher's mean squared error (MSE) 
$\E{}{ \| \pStar( \X ) - \pTeach( \X ) \|_2^2 }$, which by Proposition~\ref{prop:bias-variance} bounds the gap between the population and distilled empirical risk.

\begin{figure}[!h]
    \centering
    
    \includegraphics[scale=0.47]{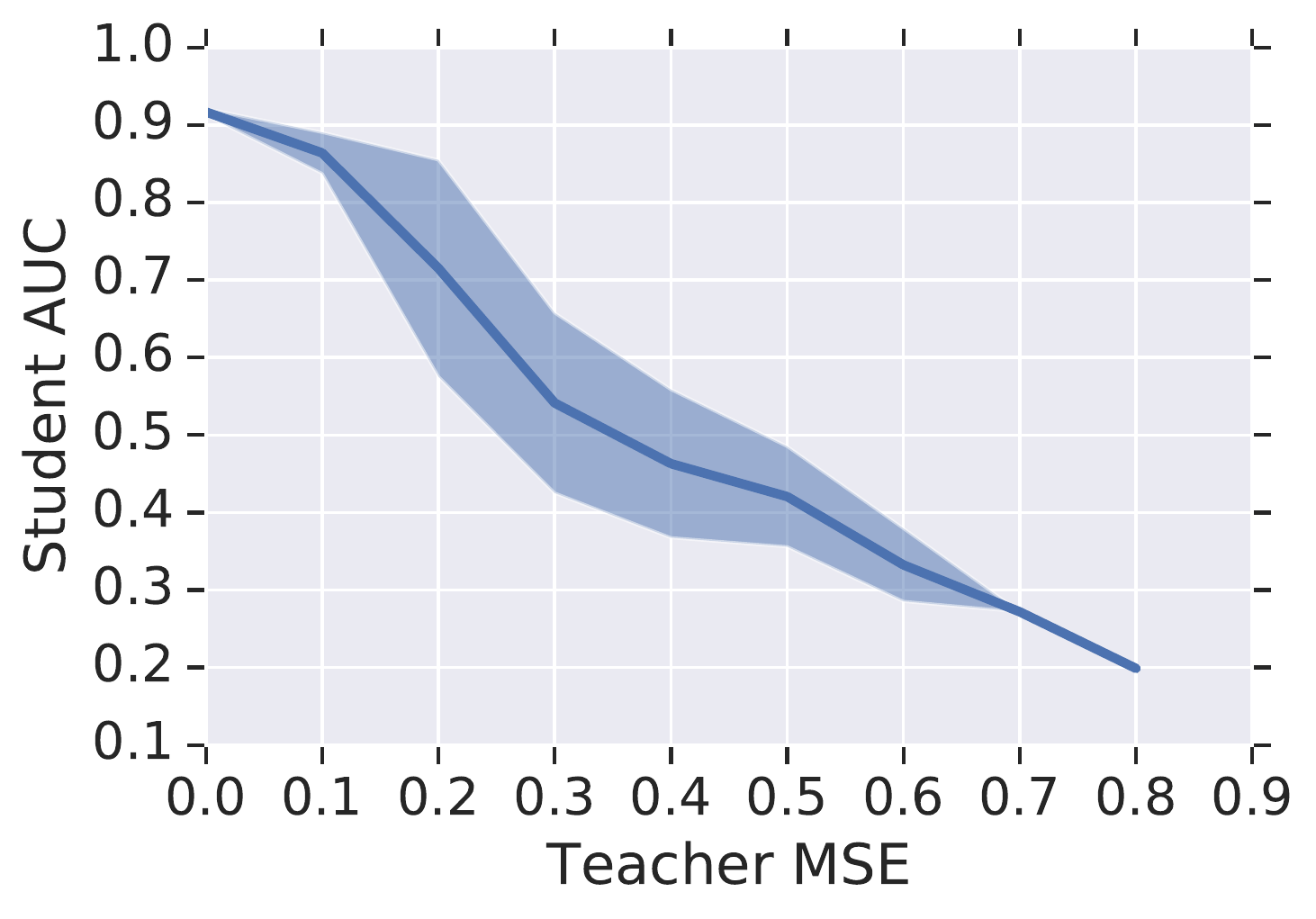}

    \caption{Relationship between teacher's MSE against true class-probability and student's test set AUC.
    In keeping with the theory,
    teachers which better approximate $\pStar$ yield better students.}
    \label{fig:gaussian_distillation_bias_variance}
\end{figure}

\begin{figure*}[!h]
    \centering
    
    \includegraphics[scale=0.5]{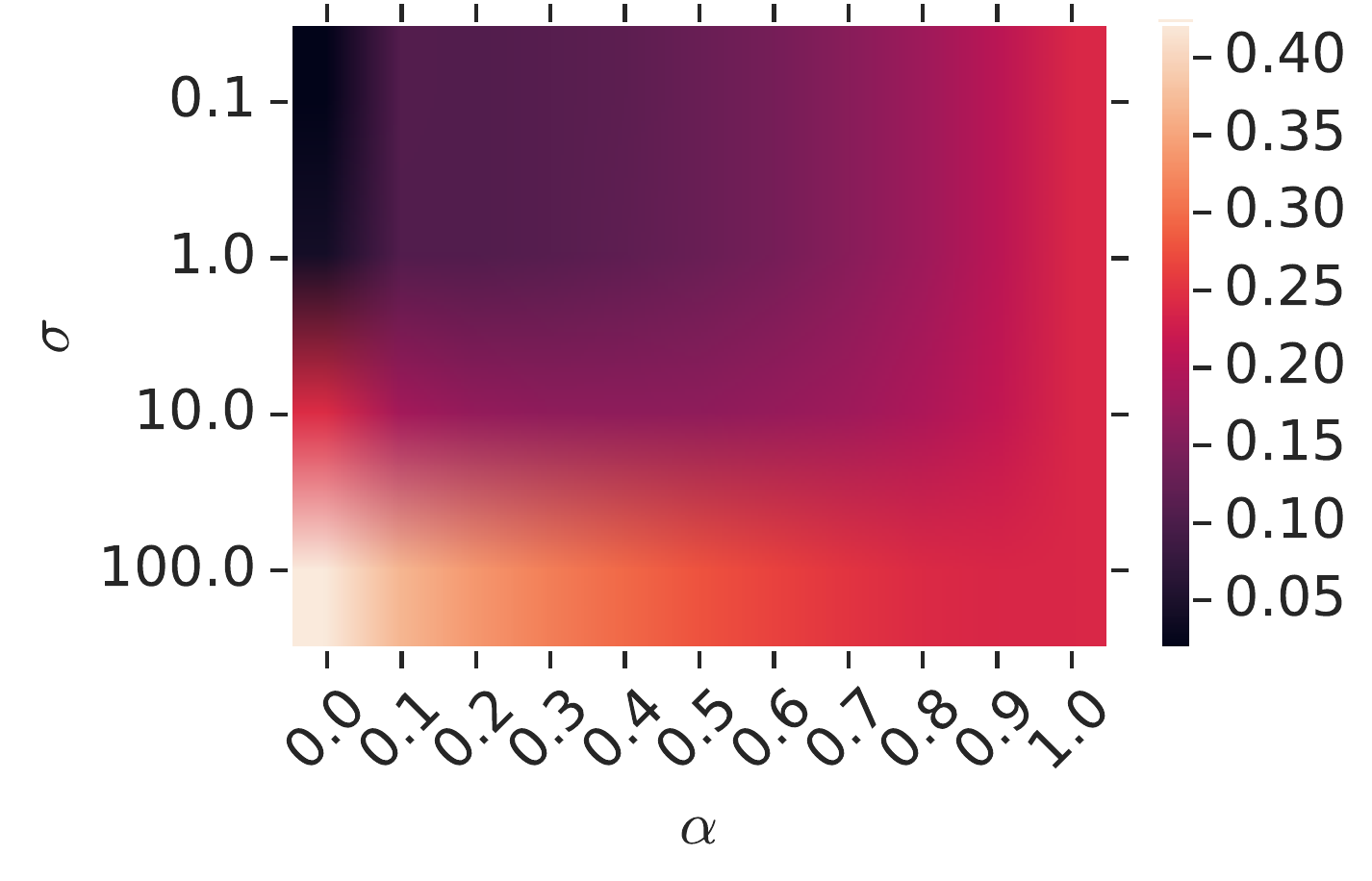}
    
    \caption{Relationship between teacher's bias and variance,
    and corresponding MSE against true class-probability.
    The teacher can achieve a given MSE through multiple possible bias-variance combinations.}
    \label{fig:gaussian_distillation_bias_variance_mse}
\end{figure*}

For each such teacher, we compute its MSE,
as well as the test set AUC of the corresponding distilled student.
Figure~\ref{fig:gaussian_distillation_bias_variance} shows the relationship between the the teacher's MSE and the student's AUC.
In line with the theory, more accurate estimates of $\pStar$ result in better students.
Figure~\ref{fig:gaussian_distillation_bias_variance_mse} also shows how the teacher's MSE depends on the choice of $\sigma$ and $\alpha$,
demonstrating that multiple such pairs can achieve a similar MSE.
As before, we see that a teacher may trade-off bias for variance in order to achieve a low MSE.


%

\subsection{{Uncalibrated} teachers may distill worse}
\label{sec:distorted_bayes_distillation}

We now illustrate the importance of the teacher probabilities needing to be meaningful reflections of the true $\pStar( x )$.
We continue our exploration of the synthetic Gaussian problem,
where $\pStar$ takes on a sigmoid form.
We now distort these probabilities as follows:
for $\alpha \geq 1$, we construct
\begin{align*}
    \bar{p}( x ) &= \Psi_{\alpha}( \pStar( x ) ) \\
    \Psi_{\alpha}( u ) &= 
    \begin{cases}
\frac{1}{2} \cdot (2 \cdot u)^{\alpha} & \text{ if } u \leq \frac{1}{2} \\
\frac{1}{2} + (2 \cdot u - 1)^{1/\alpha} & \text{ if } u \geq \frac{1}{2}.
\end{cases}
\end{align*}
The new class-probability function $\bar{p}( x )$ preserves the classification boundary $\pStar( x ) = \frac{1}{2}$,
but squashes the probabilities themselves as $\alpha$ gets larger.
We now consider using $\bar{p}( x )$ as teacher probabilities to distill to a student.
This teacher has the \emph{same accuracy, but significantly worse calibration} than the Bayes-teacher using $\pStar( x )$.

\begin{figure*}[!h]
    \centering

    \subfigure[As tuning parameter $\alpha$ is increased, the teacher probabilities $\bar{p}( x ) = \Psi_{\alpha}( \pStar( x ) )$ increasingly deviate from the Bayes probabilities $\pStar( x )$.]
    {
    \includegraphics[scale=0.45]{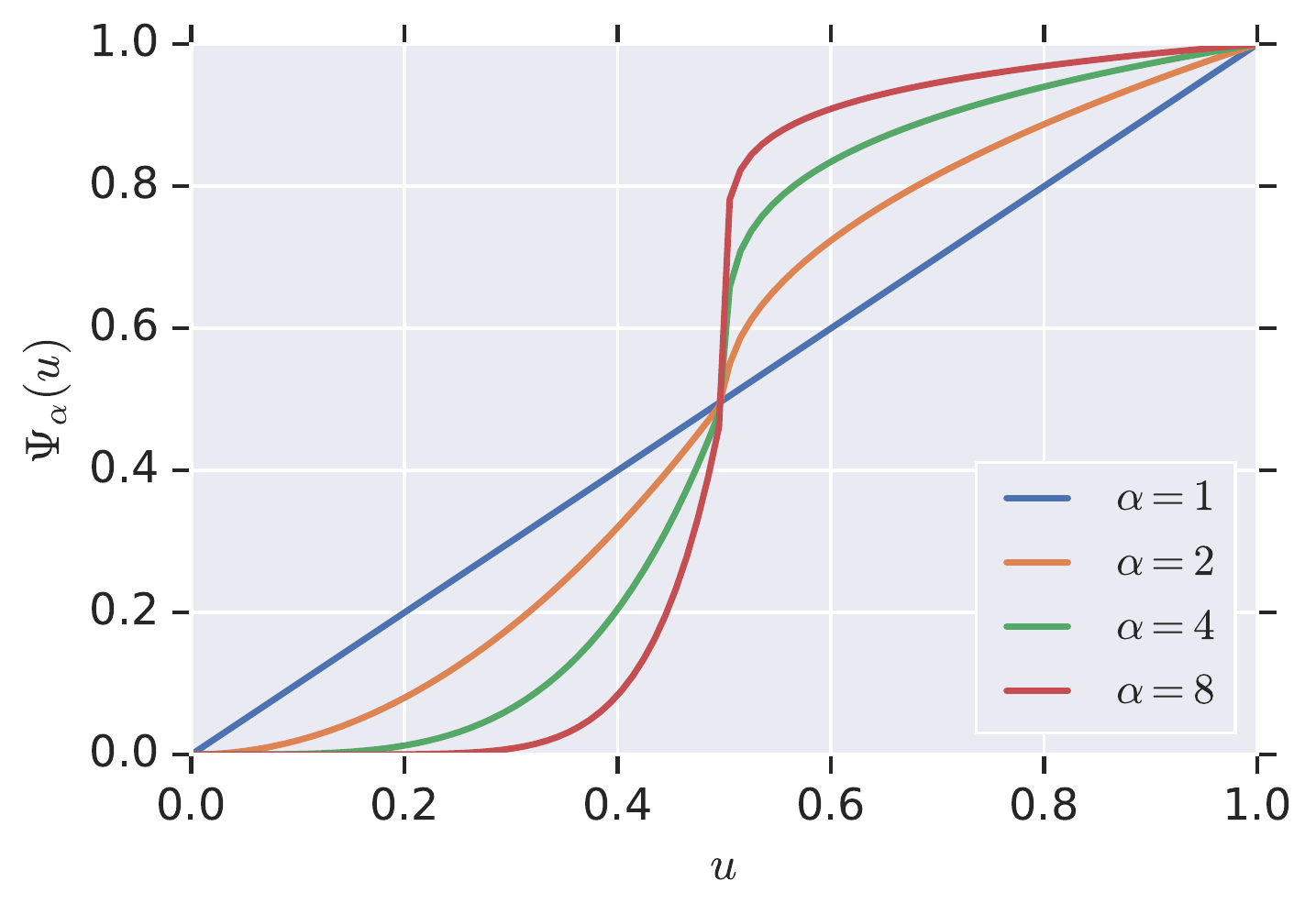}
    }%
    \qquad
    \subfigure[As tuning parameter $\alpha$ is increased ---
    so that the teacher probabilities are increasingly uncalibrated
    ---
    the student AUC becomes progressively worse.]
    {
    \includegraphics[scale=0.45]{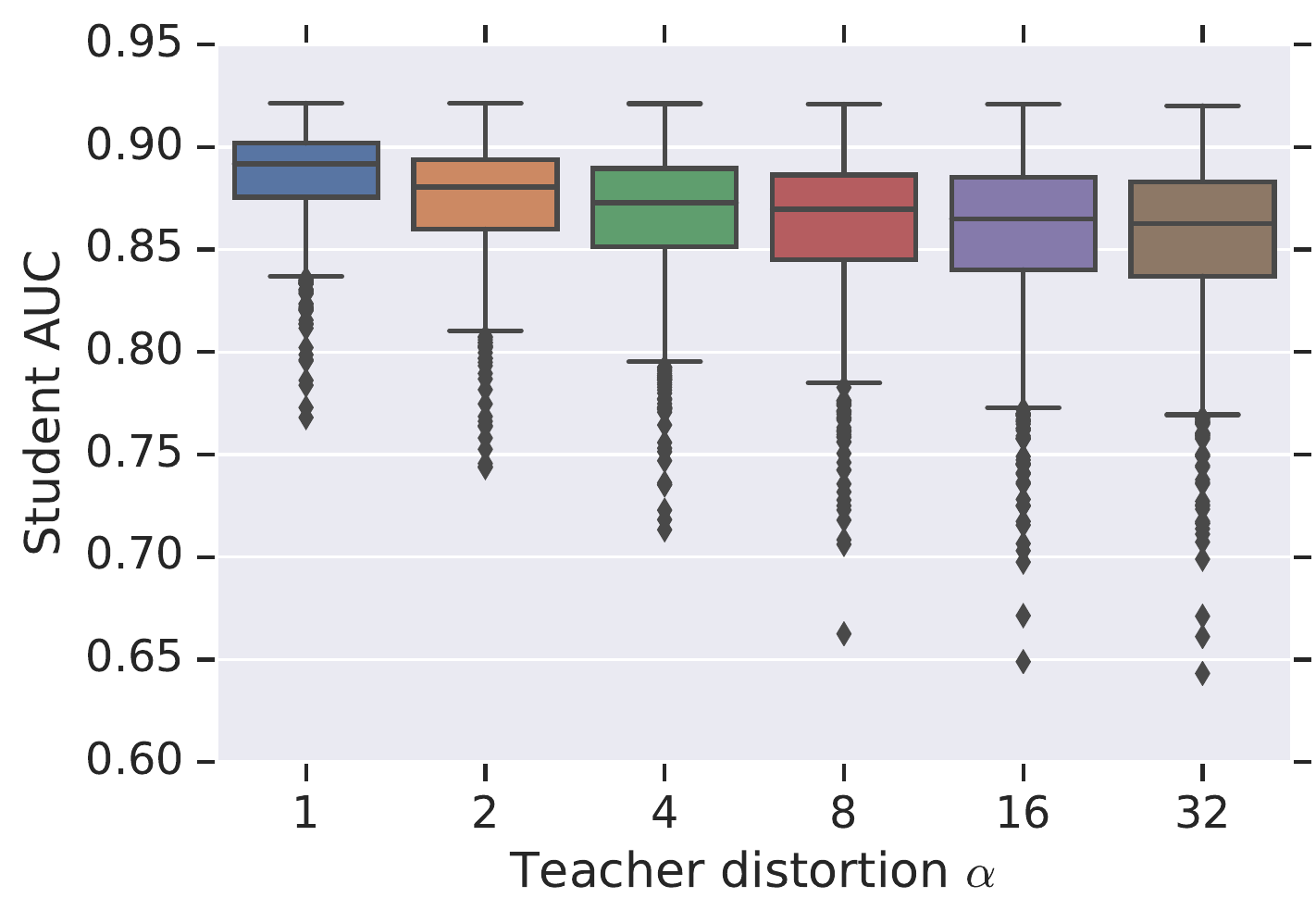}
    }%
    
    \caption{Effect of distorting the Bayes probabilities $\pStar( x )$ so as to preserve the classification decision boundary ($\pStar( x ) = \frac{1}{2}$), but degrade the calibration of the scores.}
    \label{fig:gaussian_distillation_bayes_distortion}
\end{figure*}

Figure~\ref{fig:gaussian_distillation_bayes_distortion}
confirms that as $\alpha$ increases,
the effect of distilling on the student is harmed.
This validates our claim that teacher accuracy is insufficient to judge whether distillation will be useful.

\end{document}